\documentclass[11pt]{article}
\usepackage[margin=0.9in]{geometry}
\usepackage[T1]{fontenc}
\usepackage{lmodern,microtype,amsmath,amssymb,booktabs,tabularx,graphicx,xcolor}
\usepackage{tikz,pgfplots}
\usetikzlibrary{arrows.meta,positioning}
\pgfplotsset{compat=1.18}
\usepackage{xurl}
\usepackage{hyperref}
\hypersetup{colorlinks=true,linkcolor=blue!50!black,citecolor=blue!50!black,urlcolor=blue!50!black,pdftitle={When Every Choice Is Correct: A Causal Study of Numerical Preference in Language Models}}
\usepackage{enumitem}
\usepackage{needspace,placeins}
\setlist{nosep}
\newcommand{\KL}{\operatorname{KL}}
\newcommand{\JS}{\operatorname{JS}}
\newcommand{\E}{\mathbb{E}}
\title{When Every Choice Is Correct: A Causal Study of Numerical Preference in Language Models}
\author{Hongnian Huang\\Department of Mathematics and Statistics\\University of New Mexico\\\texttt{hnhuang@gmail.com}\and Hongliang Liu\\Independent researcher\\\texttt{phunter.lau@gmail.com}}
\date{}
\begin{document}
\maketitle
\begin{abstract}
When a language model chooses a number between 1 and 30, every integer in the range is valid, yet the model can favor particular answers. We use numerical choice to study how preference is processed across layers and tokens. Perturbation probing and gradient attribution nominate neurons for activation interventions. In a selected Qwen3-4B case, a pre-answer edit changes 17 to 15 while both runs first emit 1. After the edit ends, selective cache and attention-output replacements trace part of its continuing influence on the second digit. Replacing one attention head's output or the other heads' outputs jointly produces similar immediate readout contributions; both interventions restore 17 but have different final log-odds effects on 7 versus 5. Holding later multilayer perceptron (MLP) outputs at their values from the edited run without either replacement reduces this effect gap from 0.119 to 0.0025 nat. Later computation therefore changes how similar local contributions affect the continuation. Retrospective estimates across eighteen contexts capture much of the probability movement but miss the selected answer switch. Prompt tests also separate influence from answer changes: one held-out prompt retains influence without switching, and only one of twelve fresh wordings qualifies for the reader/MLP test; there, joint MLP clamping again reduces the gap between the two reader effects while the answer remains 17. These results use numerical choice as a controlled case for studying persistent state and conditional downstream processing; they do not establish the origin or a complete circuit of natural numerical preference.
\end{abstract}

\section{Introduction}
Consider the prompt ``Give me a number between 1 and 30''. Choosing 17 satisfies this request. So do the other twenty-nine integers in the range. What determines which one a language model favors? Correctness alone leaves this choice unresolved. The same question arises when an agent can use functionally equivalent tools or a programming request leaves the implementation language unspecified. Studies report systematic preferences in tool selection and programming-language choice \cite{tools,languages}. In practical tasks, however, utility, familiarity and presentation can differ across alternatives, making the source of a preference difficult to isolate.

We use numerical choice as a controlled model system for preference among equally valid answers. Within a finite set of candidate integers, validity is easy to check, the alternatives can be enumerated, and short outputs allow us to examine generation token by token. An intervention can change the chosen answer while preserving its validity. We use \emph{preference} to mean the model's context-dependent allocation of probability among these answers. Equal validity does not require equal probability. The task reduces application-specific differences while retaining learned associations, tokenization effects and context sensitivity. These remaining influences are part of the system we study.

\begin{figure}[t]
\centering% Conceptual overview grounded in the selected Qwen3-4B continuation case.
% Control arrows denote possible influence, not exclusive edges or probabilities.
\begin{tikzpicture}[>=Latex,
  every node/.style={font=\small},
  choice/.style={draw=blue!55!black,fill=blue!3,rounded corners=3pt,
    minimum width=2.65cm,minimum height=1.05cm,align=center},
  control/.style={draw=gray!45,fill=gray!3,rounded corners=3pt,
    text width=2.65cm,minimum height=.95cm,align=center,font=\footnotesize},
  flow/.style={->,line width=1.05pt,draw=blue!60!black},
  influence/.style={->,line width=.75pt,draw=teal!70!black},
  carry/.style={->,line width=.65pt,draw=gray!65}]
\node[align=center,text width=2.1cm] (prompt) at (0,0)
  {Pick a number\\1--30};
\node[choice] (first) at (3.7,0)
  {{\Large\bfseries 1}\quad {\color{gray}2\quad3\quad$\cdots$}};
\node[choice] (second) at (8.0,0)
  {{\Large\bfseries 7}\quad {\color{gray}5\quad$\cdots$}};
\node[draw=blue!60!black,fill=blue!8,rounded corners=3pt,
  minimum width=1.5cm,minimum height=1.05cm,font=\LARGE\bfseries]
  (answer) at (11.6,0) {17};
\node[font=\footnotesize] at (3.7,1.0) {First-token competition};
\node[font=\footnotesize] at (8.0,1.0) {Next-token competition, given \textbf{1}};
\node[font=\footnotesize] at (11.6,1.0) {Answer};
\draw[flow] (prompt)--(first);
\draw[flow] (first)--node[above,font=\footnotesize]{emit \textbf{1}}(second);
\draw[flow] (second)--node[above,font=\footnotesize]{emit \textbf{7}}(answer);

\node[control] (context) at (0,-2.1) {Prompt and context};
\node[control] (source) at (3.7,-2.1) {Source MLP activation};
\node[control] (memory) at (8.0,-2.1) {Cached state\\\mbox{Attention readers}};
\node[control] (late) at (11.6,-2.1) {Later MLP responses};
\draw[influence] (context.north) to[out=90,in=230] (first.south west);
\draw[influence] (source.north)--(first.south);
\draw[carry] (source.east)--(memory.west);
\draw[carry] (memory.east)--(late.west);
\draw[influence] (memory.north)--(second.south);
\draw[influence] (late.north) to[out=90,in=310] (second.south east);
\node[font=\scriptsize,text=gray!75!black,align=center,text width=12.7cm]
  at (5.8,-3.15)
  {Schematic control sites; effects can reinforce or oppose a choice.};
\end{tikzpicture}
\end{figure}

Numerical-generation studies report failures under explicit uniform or independent sampling requirements \cite{numbers,stochastic,renda,west}. Those requirements go beyond choosing a valid number. Recent matched-checkpoint comparisons show that instruction tuning changes number-choice distributions without producing one shared favorite integer across models \cite{ansari}. In a related two-option setting without a correct answer, Baroni et al.\ \cite{entropy} study instructed entropy changes and activation steering. Digit-selective neurons and selective digit interventions also have precedents in numerical reasoning \cite{benford,modular,base10}. These studies motivate examining both the distribution of choices and the computation that changes it. We ask how an activation edit redistributes preference, how its influence persists across tokens, and how later computation transforms that influence.

We combine perturbation probing (PP) and gradient nomination with scalar patches and zeroing to select a source neuron. PP supplies forward candidate scores \cite{pp}; gradient-based tracing in the neuron basis identifies causally effective components and supports group and individual-neuron steering \cite{sparse}. Downstream compensation \cite{hydra,selfrepair} and dose-dependent control limits \cite{control,strength} motivate following the edit beyond its initial effect. A successful intervention identifies a control point but does not by itself establish how the altered influence is processed, or how much natural behavior it explains \cite{vig,patching,illusion}. We therefore connect source interventions to probability estimates, selective downstream replacements and natural-context transfer tests.

A selected Qwen3-4B example makes the processing question concrete. Increasing one multilayer perceptron (MLP) activation at the pre-answer position changes the generated answer from 17 to 15, while both runs first emit 1. The edit ends before the model processes that shared digit, so the visible prefix is unchanged while inherited internal state can differ. Selective source-position cache and attention-output replacements trace part of the continuing influence (Figure~\ref{fig:overview}). Two attention-output replacements have similar immediate readout contributions but different final effects. Holding later MLP outputs at their values from the edited run without a reader replacement largely removes this difference. The comparison connects persistent state to conditional downstream processing.

Our contributions are:
\begin{itemize}
\item \textbf{Partial propagation and downstream processing.} Selective cache and reader replacements localize part of a pre-answer-only 17-to-15 edit across an unchanged first digit. Later-MLP output clamping changes the relative reader effects while both restored answers remain 17.
\item \textbf{Probability movement and answer changes.} A frozen two-target panel tests signed effects against matched-write controls. A retrospective shared-profile estimate captures much of the probability movement across eighteen contexts but misses the selected near-boundary switch; matched controls qualify the response-shape specificity.
\item \textbf{Steering and natural-context transfer.} Natural-activation swaps test how much instruction and history variation the steering target carries. Complement, held-out and fresh-prompt tests separately bound the specificity and portability of the selected pathway.
\end{itemize}
The central saved-record analyses and continuation study concern Qwen3-4B; earlier studies in three related checkpoints provide supporting context. On one new prompt, the same source and head retain signed effects while all tested continuations remain 17. The fresh twelve-wording screen adds one eligible reader/MLP recurrence; the other eleven wordings have no downstream intervention evidence. These tests bound reuse of the selected components. The selected path supplies a conditional explanation of an imposed preference change. The training origin of natural numerical preference and a minimal circuit for numerical choice remain open.

\section{Experimental setup and causal interventions}
\subsection{Models, task and answer measurements}
We study choice within a finite set of candidate integers. A validity check determines whether an answer satisfies the range constraint; the relative probabilities describe preference within that set. We use scored distributions and sampled or greedy replies for different purposes. A repeated greedy answer alone does not establish probability concentration.

We studied Qwen3-4B, Qwen3-4B-Instruct-2507 and Qwen3-8B \cite{qwen}. These checkpoints share an architecture family; 2507 is a related variant. The selected MLP neurons are respectively L29:8172, L29:8172 and L29:385, with zero-based indices. A five-checkpoint search also included Qwen3-14B and GPT-OSS-20B, where the single-unit criterion failed. Supporting response tests concern these three checkpoints; the detailed continuation pathway uses Qwen3-4B.

Let $x$ be the prompt and $h$ a fixed response prefix. For candidates in interval $I$, define
\begin{equation}
 w_h(n)=P(\text{canonical digits of }n\text{ followed by a nondigit}\mid x,h),
 \quad q_h(n)=\frac{w_h(n)}{\sum_{k\in I}w_h(k)}.
\end{equation}
The denominator is the mass of this candidate-event set. This event admits a decimal point as a terminator, and is therefore not exactly the probability of an integer-only completed answer. Number words, alternative spellings and signs are outside this definition. Ansari \cite{ansari} separately reports admitted answer mass and conditional number-choice probabilities using a different spelling and termination rule. Our results using these scores are explicitly separated from parsed generated replies; the two event definitions are not interchangeable.

An own-prefix evaluation retains the model's natural text before the answer. Other experiments use the fixed prefix ``The number is ''. These interfaces condition the measured distribution on different histories, and a change of prefix can change the result. Generated-answer checks therefore accompany the scored-event analyses where available.

The baseline preference differs across these checkpoints. In the earlier 1--30 own-prefix scores, $q(17)$ was 0.79 in Qwen3-4B and 0.82 in Qwen3-4B-Instruct-2507. Qwen3-8B assigned 0.53 to 15 and 0.47 to 17. These conditional candidate scores illustrate unequal preference without a shared winner across all three checkpoints; they are not frequencies of unrestricted replies. They also concern a different cohort from the selected switching prompt in Section~\ref{sec:path}.

We saved completed replies where generated-answer checks were available. In the selected dose study, we compared full-chat generation with continuation from a retained prefix. This checks their agreement in the tested conditions; it does not identify candidate-event probabilities with unrestricted reply frequencies.

\subsection{Candidate nomination and causal selection}
An MLP neuron is an intermediate activation $z$ immediately before the down projection. It contributes $zv$ to the residual stream, where $v$ is the corresponding weight column. Decomposing MLP outputs into such writes and examining their vocabulary-space effects follows established analyses \cite{ffn}. Our interventions change activations while keeping model weights fixed.

For Qwen3-4B, we compared twelve prompt templates at 1--30 with donor prompts at 1--40, 1--50, 1--35 and 1--20. Each prompt supplied its own generated pre-answer prefix, and the donor's chosen number supplied the competing answer. We measured activation displacements at these numerical decision positions.

Two branches nominated candidates. The forward perturbation-probing (PP) score multiplies the displacement by the projection of the down-projection column onto the first-digit unembedding contrast \cite{pp}. The gradient-attribution score multiplies the displacement by the gradient of the complete-numeral event log-margin between 17 and the donor answer. This first-order screening follows the attribution-patching approach studied by Kram\'ar et al.\ \cite{atp}; we do not implement their AtP* corrections. Their across-pair magnitude aggregation also differs from the signed-mean PP aggregation below. The branches therefore assess different readouts: first-digit coupling and the numeral event defined above. For each range edit $e$, let $T_e$ be the set of eligible templates and $u_{nte}$ the signed PP score for neuron coordinate $n$ and template $t\in T_e$. The corrected aggregation is
\begin{equation}
 s_{ne}=\left|\frac{1}{|T_e|}\sum_{t\in T_e}u_{nte}\right|.
\end{equation}
The sign is retained for interpretation. We normalized and combined scores across range edits, ranked coordinates in the last quarter of decoder layers, and pooled the top twenty coordinates from each branch, reranked their union by the summed normalized scores, and retained twenty candidates. Appendix~\ref{app:cohorts} specifies the normalization and selection details. The layer window restricts the search; it is not evidence that earlier layers are uninvolved.

We then measured finite effects with single-scalar patches and zeroing. We retained candidates whose zeroing preserved the pooled baseline modes at 1--10, 1--30 and 1--100, including 17 at 1--30. Among eligible candidates, we selected the coordinate with the largest increase in $q(17)$ at 1--40 under the frozen rule. Exploratory ablations had already identified the leading candidate before this frozen comparison. This selection preceded held-out evaluation; 1--40 remains a selection interval. Later additive-dose and path tests characterize this selected source under their own intervention conditions.

In the corrected Qwen3-4B search, L29:8172 ranks first in the gradient branch and 31st in the PP branch, outside its top twenty. It therefore enters through gradient attribution and remains the intervention-screen winner. The primary correction changes PP aggregation; a separate sensitivity analysis changes both branches. The 2507 and 8B coordinates were selected under the historical magnitude-aggregation procedure; corrected nomination was not repeated for them (Appendix~\ref{app:evidence}).

\subsection{Interventions and comparisons}
We use \emph{registered} for predictions and criteria recorded internally before the corresponding run, without a public registration. Exploratory selection, reused cohorts and post hoc analyses are identified separately. Setting, zeroing or adding to the source activation changes its residual write. Activation patching replaces a recorded internal value with the corresponding value from another run. Scalar donor swaps use a value from a different prompt; residual and cache interventions replace larger objects. Response blocking holds specified later module outputs at their recipient baseline when testing a source or reader intervention. Context already encoded in those baseline outputs remains available. Table~\ref{tab:operations} summarizes the interventions in the selected continuation study.

We distinguish an imposed change within a prompt from transfer of a natural contrast between prompts. Each comparison specifies its intervention position, recipient background and readout. First-digit, candidate-answer and fixed-leading-digit distributions answer different questions about generation. We retain these distinctions and matched controls when reporting effects \cite{patching}. Circuit-level evaluations distinguish faithfulness, completeness and minimality \cite{ioi}; our selective replacements identify partial contributions without establishing those properties for a complete circuit. Specialized metrics are introduced with the corresponding results.

\begin{table}[htbp]
\centering\small
\begin{tabularx}{\linewidth}{>{\raggedright\arraybackslash}p{2.5cm}X}
\toprule
Intervention & Changed object and reference \\
\midrule
Source dose & Add $\delta$ to the selected pre-down-projection scalar at the last pre-answer position; remove the hook before processing the first digit. \\
Cache replacement & Copy selected source-position keys or values from the stated donor cache into the recipient. Other entries remain unchanged. \\
Reader replacement & Copy selected L33 head-output slices before the attention output projection, while processing the common first digit. \\
Late-MLP clamp & Replace L34--35 MLP output vectors at that query with their source-edited, no-reader values. Attention and final normalization remain active. \\
\bottomrule
\end{tabularx}
\caption{Operations in the selected continuation study. All model weights remain fixed. Cache and reader donors are specified with each comparison; replacement directions do not prescribe a particular sign or answer. Source zeroing in the earlier assays has its separately stated position scope.}
\label{tab:operations}
\end{table}

The principal prediction and interval-response results used FP32 confirmation, and the later pathway study used unquantized FP32 with streamed CUDA. The experiments were run on the RTX 2070 Max-Q host reported by the authors; Appendix~\ref{app:evidence} distinguishes device records, selection history and runtime coverage.

\section{Structured changes in numerical preference}\label{sec:prediction}
\subsection{Neuron edits retain their effect on new wordings}
We first tested whether a local edit could redirect numerical preference beyond the wording used to select the source. The integer 17 remains a valid answer throughout this comparison. At 1--40, zeroing the selected neuron increased sampled 17 replies in all three checkpoints (Appendix~\ref{app:historical-prediction}; Figure~\ref{fig:foundation}A). The new-wording cohort contains twenty paraphrases, of which 16, 17 and 20 meet the eligibility rule. Each eligible prompt supplies 32 sampled replies after its own preamble, with temperature 0.7, top-$p$ 0.8 and top-$k$ 20. The fraction of sampled replies equal to 17 increased from 0.02 to 0.77 in Qwen3-4B, from 0.00 to 0.69 in Qwen3-4B-Instruct-2507, and from 0.15 to 0.95 in Qwen3-8B. The corresponding 95\% template-bootstrap intervals for the \emph{increase} are [0.56,0.92], [0.47,0.88] and [0.66,0.92].

Appendix~\ref{app:cohorts} reports the reused-template cohort separately. Free generation with zeroing at every position also increased 17 replies, although that assay changes both the intervention scope and the response-prefix treatment. Because the 1--40 interval participated in selection, the new-wording result tests transfer across eligible phrasings at that interval. The historical predictor addressed new intervals in a separate study.

One hundred random neurons matched on down-projection-column norm produced maximum $q(17)$ increases of 0.003, 0.000 and 0.001 (rounded source summaries). Matching a weight norm does not match the activation-dependent write removed by zeroing. A separate later comparison uses one matched-write coordinate; it answers a narrower control question. Protected answers in the original arithmetic, copy and comparison checks remained correct, but no checkpoint met every registered validation criterion. A larger arithmetic study reports small mean probability changes with a nontrivial sensitive tail (Appendix~\ref{app:arithmetic}). These checks constrain specificity without proving that the neuron is uniquely numerical.

\subsection{Frozen doses test directional response}\label{sec:fixed-dose}
We next tested whether fixed doses retain a predicted direction across contexts. This Qwen3-4B panel used twelve historical templates (P1--P12) and six extensions (E1--E6), each with its own pre-answer prefix at 1--30. L29:8172 and a separately selected L34:2836 were edited individually, with one residual-write-matched control per site. The protocol predicted that a positive dose would favor 17 over 23 at L29 and oppose it at L34; negative doses should reverse those directions. The endpoint was the change in $\log w(17)-\log w(23)$, with effects of magnitude at most 0.05 nat classified as unresolved. This floor is a practical reporting criterion, not a significance test. Appendix~\ref{app:fixed-protocol} specifies target selection, exact doses, control matching and all eighteen prompts, with a link to the frozen predictions.

\begin{table}[ht]
\centering\small\begin{tabular}{llrrr}
\toprule
Target & Cohort & Predicted & Unresolved & Mean $|\Delta|$ (control)\\
\midrule
L29:8172 & Historical & 24/24 & 0 & 0.636 (0.060)\\
L29:8172 & Extensions & 10/12 & 2 & 0.533 (0.065)\\
L34:2836 & Historical & 24/24 & 0 & 0.317 (0.007)\\
L34:2836 & Extensions & 12/12 & 0 & 0.357 (0.004)\\
\bottomrule
\end{tabular}

\caption{Frozen directional predictions at two individually edited sites. Each prompt supplies positive and negative target doses. Parentheses give mean absolute effects for the site's matched control; effects are in nats. Historical contexts informed selection; the six extensions are outside that discovery set. These dependent conditions do not estimate a population success rate.}
\label{tab:fixed-signs}
\end{table}

Seventy of 72 target doses met the predicted-direction criterion, including 22 of 24 on the extensions; none had a resolved opposite effect. The two unresolved L29 doses occur in E2, and its positive dose has a small opposite raw sign. L29's control meets the target's directional prediction on 20 of 36 doses, is unresolved on fourteen and is opposite on two; the L34 control is unresolved on all 36. The selected L29 source produced larger effects than this control, which also frequently met the directional criterion. These are separately tested sites, not evidence of a connection between them. Appendix~\ref{app:cohorts} retains the full panel accounting and donor qualification.

The directional test concerns one rival, not the winning answer. We use the same saved distributions below to examine all thirty alternatives. The 17/15 analysis and selection of the switching prompt were post hoc. Earlier studies separately report leading-digit prediction across new intervals, but their full fit and evaluation records are unavailable; Appendix~\ref{app:historical-prediction} preserves those qualified summaries rather than treating them as a new validation.

\subsection{Shared responses predict distribution changes}
The directional test evaluates one rival. We next ask whether a shared response can estimate the redistribution over all thirty candidate answers.

To compare the responses directly, we examined changes in per-number log probabilities at the same source dose. Leading 2 loses against leading 1 in all eighteen contexts, with median log-odds change $-0.651$ nat. The 17/15 contrast has a median change of $-0.282$ nat and decreases in all eighteen contexts; the control's median change is $+0.061$ nat. Sixteen response profiles correlate at least 0.96 with the coordinate-wise median of the other seventeen profiles across all thirty candidates, and at least 0.92 within 10--19. The two less similar profiles begin with candidate modes 23 and 15 (Figure~\ref{fig:response-geometry}A). This is a retrospective description of a related template family; equal-weight correlations also count extremely unlikely answers.

We evaluated these profiles as retrospective distribution estimates: each prompt receives the median response from the other seventeen, applied to its own baseline and renormalized. This reduces pooled Kullback--Leibler (KL) prediction error relative to the unedited baseline by 89.0\%, with a median per-prompt reduction of 96.3\%; all eighteen estimates improve. Appendix~\ref{app:profile-prediction} reports absolute losses, all contexts and cohort summaries. These related contexts provide a retrospective diagnostic, not prospective validation of a frozen predictor.

A controller that only changes leading-digit weights has a structural limit: within a set such as 11--19, every candidate receives the same leading-digit multiplier, which cancels on normalization. It predicts no redistribution inside that set. The earlier FP32 clamp study nevertheless reports within-set total variation (TV) changes of 0.06--0.17. This documented summary concerns the interval study, not the later selected dose. It motivates asking how an edit changes the continuation after a shared first digit.

We can measure both kinds of redistribution exactly in the later saved distributions. Let $q_a$ and $q_0$ denote the candidate-answer distributions with intervention $a$ and at baseline, respectively, and let $d(n)$ be the leading digit of candidate $n$. The leading-digit mass and within-digit conditional distribution are $p_a(d)=\sum_{n:d(n)=d}q_a(n)$ and $r_a(n\mid d)=q_a(n)/p_a(d)$. The KL chain rule gives
\begin{equation}
 \KL(q_a\Vert q_0)=\underbrace{\KL(p_a\Vert p_0)}_{B_a}
 +\underbrace{\sum_d p_a(d)\KL(r_a(\cdot\mid d)\Vert r_0(\cdot\mid d))}_{C_a}.
 \label{eq:branch-continuation}
\end{equation}
A branch oracle that uses observed $p_a$ but retains $r_0$ leaves error $C_a$. Thus even perfect knowledge of the changed branch weights cannot explain within-branch redistribution. This is retrospective accounting, with no fitted coefficients or new model calls.

A branch-only median comparator gives 38.5\% pooled reduction. Even an oracle supplying the observed branch masses while retaining baseline within-branch probabilities is limited to $R^\star_{\mathrm{branch}}=\sum_j B_j/\sum_j(B_j+C_j)=40.9\%$ over the eighteen contexts. The full-profile advantage therefore exceeds what any predictor in this branch-only class can achieve; it is not explained by an ineffective branch estimator. This output-level bound does not identify an internal carrier.

The matched-write control also has a consistent response shape within 10--19, with a smaller amplitude and opposing alignment to the source. The shared shape therefore describes output structure without identifying a source-specific signature or a common internal route. Appendix~\ref{app:profile-prediction} gives the complete comparator and sensitivity analyses.

\subsection{From probability estimates to answer changes}
The selected prompt illustrates why distributional accuracy and winner prediction differ (Figure~\ref{fig:response-geometry}B). Its estimate reduces KL error by 88.9\% but retains 17: the estimated 17/15 margin is $+0.110$ nat, whereas the observed margin becomes negative. The selected response has a familiar shape but enough displacement to cross a near tie. Selection on the observed switch makes this an existence example. Baseline and estimated modes both agree with the observed mode in seventeen of eighteen contexts, so mode accuracy alone adds little here.

\begin{figure}[tb]
\centering\input{figures/response_geometry}
\caption{A shared response captures probability movement while missing the selected switch. A: all eighteen source responses at the same dose, centered across the thirty candidate log probabilities; the black curve is their coordinate-wise median. E1 and E2, initially favoring 23 and 15, remain visible. B: the selected prompt's baseline, leave-one-out estimate from the other seventeen responses, and observed source distribution. The estimate retains mode 17; the observed source favors 15. For the full thirty-candidate distribution in B, KL prediction error falls by 88.9\%, although the estimated mode remains 17. E1 uniquely has no pre-answer text after the chat template. This retrospective comparison is distinct from the earlier fitted interval predictor.}
\label{fig:response-geometry}
\end{figure}

The prediction miss motivates a closer test of the continuation computation. Both candidate answers begin with 1, so this selected case lets us follow an edit's influence into a later digit without changing the first emitted token.

\FloatBarrier
\Needspace{12\baselineskip}
\section{An unchanged first digit reveals a continuation pathway}\label{sec:path}
We first examine a prompt selected for an observed 17-to-15 switch. An earlier held-out prompt is subsequently reused for the late-MLP comparison, and a separate fresh wording screen tests recurrence under fixed eligibility rules. These studies retain their distinct selection and reuse status throughout.
\subsection{Changing the answer while preserving the first digit}
The leading-digit account leaves open how an edit changes preference within a branch. In a selected Qwen3-4B prompt, a positive edit to L29:8172 changes the completed answer from 17 to 15 while leaving the greedy first token 1 unchanged. Both answers satisfy the same 1--30 constraint. The intervention acts only at the pre-answer position; it is absent when the model processes the emitted digit. This provides a controlled setting for following an earlier activation change into a later choice.

The post hoc 17/15 comparison uses the eighteen contexts of Section~\ref{sec:fixed-dose}. The 17-versus-15 candidate-event margin decreased in all eighteen, by a mean of 0.284 nat; seventeen decreases exceeded 0.05 nat. A residual-write-matched edit to L29:8369 increased that margin in all eighteen, averaging 0.074 nat. The active control therefore supplied an opposing response under the matched-write comparison.

The frozen panel measured 199 of 216 planned conditions, with seventeen natural-donor write-cap skips. The exploratory search for an answer change then examined those measured distributions. Only one measured intervention condition changed the candidate mode relative to the corresponding prompt's baseline. This observation selected ``Pick one integer in the inclusive interval [1, 30].'' for the downstream experiments. Its baseline absolute 17/15 candidate-event margin is 0.390 nat, compared with a median of 19.775 across the eighteen prompts; the other seventeen span 3.031--22.860 nat. Only two prompts assign more than 1\% candidate probability to both 17 and 15. This near-boundary selection makes the pathway an existence example whose broader answer-switching rate remains unmeasured.

\begin{figure}[tb]
\centering\begin{tikzpicture}
\begin{axis}[width=.43\linewidth,height=5.1cm,axis lines=left,tick label style={font=\small},label style={font=\small},title style={font=\small},legend style={font=\scriptsize,draw=none},ymajorgrids,grid style={gray!20},xlabel={Dose / base dose},ylabel={Event log-odds 17/15},title={A. Selected prompt},legend pos=south west]
\addplot[orange!85!black,mark=*,thick] coordinates {(0,0.389720917) (0.25,0.2864151) (0.5,0.181999207) (0.75,0.0765380859) (0.875,0.0234222412) (1,-0.0299949646) (1.125,-0.0836830139) (1.25,-0.137561798) (1.5,-0.246143341)};
\addlegendentry{Target}
\addplot[blue!70!black,mark=*,thick] coordinates {(0,0.389720917) (0.25,0.422519684) (0.5,0.455345154) (0.75,0.488323212) (0.875,0.504894257) (1,0.521419525) (1.125,0.538028717) (1.25,0.554672182) (1.5,0.58795929)};
\addlegendentry{Control}
\addplot[black,dashed] coordinates {(0,0) (1.5,0)};

\end{axis}
\end{tikzpicture}\hfill
\begin{tikzpicture}
\begin{axis}[width=.43\linewidth,height=5.1cm,axis lines=left,tick label style={font=\small},label style={font=\small},title style={font=\small},legend style={font=\scriptsize,draw=none},ymajorgrids,grid style={gray!20},xlabel={Dose / base dose},ylabel={Candidate-event probability},ymin=.3,ymax=.7,title={B. Smooth competition},legend pos=north east]
\addplot[blue!70!black,mark=*,thick] coordinates {(0,0.596142134) (0.25,0.571047277) (0.5,0.545306111) (0.75,0.519059238) (0.875,0.505790667) (1,0.492438544) (1.125,0.479029543) (1.25,0.46560317) (1.5,0.438715337)};
\addlegendentry{q(17)}
\addplot[orange!85!black,mark=square*,thick] coordinates {(0,0.40373479) (0.25,0.428828428) (0.5,0.454568265) (0.75,0.480813725) (0.875,0.494081578) (1,0.507432975) (1.125,0.520841246) (1.25,0.534266888) (1.5,0.561153288)};
\addlegendentry{q(15)}

\end{axis}
\end{tikzpicture}
\caption{The eighteen-condition dose scan. A: source and matched-control candidate-event margins. B: the source's two leading event probabilities. Markers are measured factors; straight segments are visual guides. Actual source replies are 17 through factor 0.875 and 15 from factor 1 onward; control replies remain 17. This is one selected prompt.}
\label{fig:dose}
\end{figure}
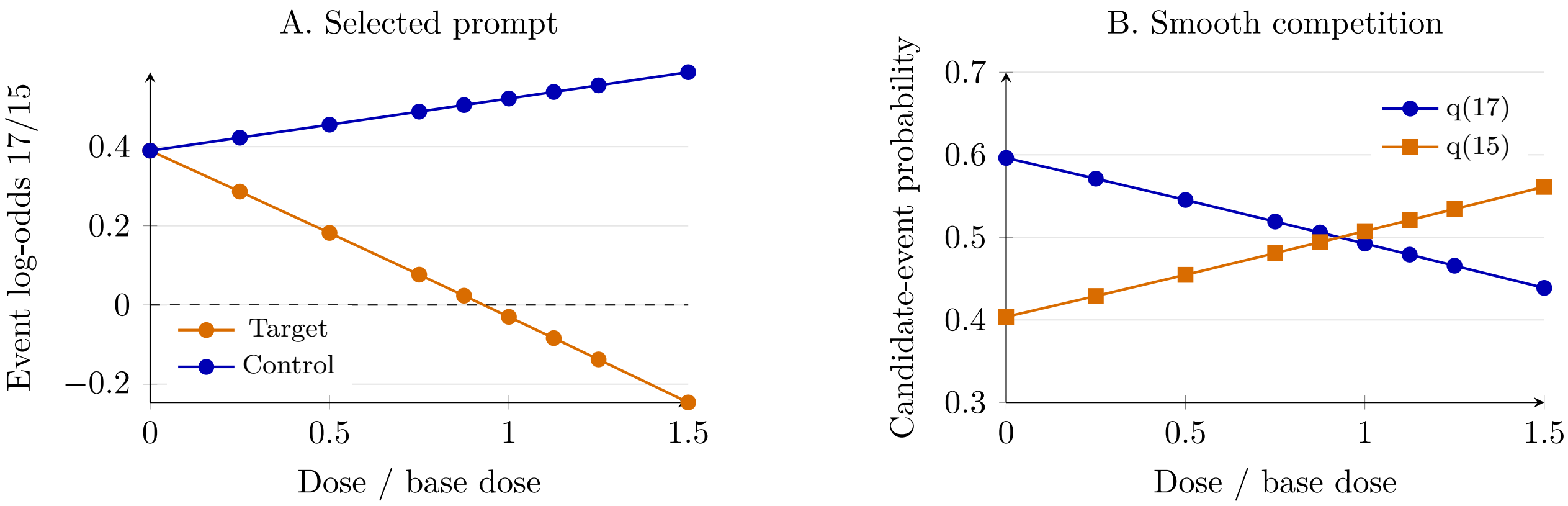

The base dose $\delta=1.81884765625$ was fixed before inference at one quarter of the median absolute displacement to the 1--35 donors in the historical template bank. The selected activation is $-9.857834$; adding the base dose places it nominally at $-8.038986$, toward zero. In the finer dose scan, both full-chat and retained-prefix replies change from 17 to 15 between factors 0.875 and 1.0 (Figure~\ref{fig:dose}). Interpolating the event-margin zero gives approximately 0.930; the actual transition was only bracketed by measured doses. A matched control retains 17 throughout the tested grid. The response is gradual even though greedy output changes discretely.

The first digit is shared, but the next-token computation differs. We measure the following log-probability contrast after processing 1:
\begin{equation}
 m(h)=\log P(7\mid x,h,1)-\log P(5\mid x,h,1).
 \label{eq:margin}
\end{equation}
The sign of this margin identifies the favored token in the 7/5 pair. Other vocabulary competitors and the complete answer require separate measurements.

In the downstream experiments, the next-token margin is $+0.389778$ without the edit and $-0.029976$ with it. Both states retain first digit 1. Candidate-event and next-token measurements agree closely on this selected context, but remain distinct endpoints. The unchanged first digit and the measured second-digit change motivate testing which cached states carry the edit forward. Detailed branch accounting appears in Appendix~\ref{app:branch-analysis}.

The selected edit also illustrates rival-dependent competition. It raises the 17-versus-23 log-odds by 0.403 nat while reducing the probability of 17 itself (Table~\ref{tab:rivals}). The reason is that 23 falls proportionally further, while 15 gains enough mass to become the candidate mode. An intervention can therefore favor 17 against one rival and disfavor it against another. Pairwise improvement alone does not identify the winning answer.

\begin{table}[ht]
\centering\small
\begin{tabular}{lrr}
\toprule
Selected retained-prefix assay & Baseline & Positive L29 edit\\
\midrule
$q(17)$ & 0.596142 & 0.492439\\
$q(15)$ & 0.403735 & 0.507433\\
$q(23)$ & $1.77260\times10^{-7}$ & $9.78779\times10^{-8}$\\
$\log[w(17)/w(23)]$ (nat) & 15.028371 & 15.431160\\
Candidate-event mode & 17 & 15\\
\bottomrule
\end{tabular}

\caption{The same dose improves one comparison while changing the winner against another competitor. Probabilities and log ratios are derived from the same full-precision event-weight vector for each condition, then rounded for display. Separately computed directional-panel margins differ by less than $5\times10^{-5}$ nat here; the 0.403-nat conclusion is unchanged. Sampled reply frequencies require a separate generation assay. The dose scan independently checks generated answers within the same experimental programme.}
\label{tab:rivals}
\end{table}

\subsection{From the pre-answer edit to source-position cached state}
The unchanged digit identifies the next causal question: which altered past states remain available when that same token is processed? Belitsky et al.\ \cite{cachesteer} show that a one-time KV-cache intervention can influence later generation. Here the cache difference is induced by an earlier scalar MLP edit; selective exchanges test which parts carry its continuation effect. In this decoder, the L29 MLP edit occurs after the block's attention computation. It can change source-position keys and values (K/V) in L30--35, while the already-computed L0--29 caches and earlier token positions remain unchanged. Saved tensors verified both locality restrictions. The source hook is removed after prefill, so the subsequent effect is not a fresh edit at the first-digit position.

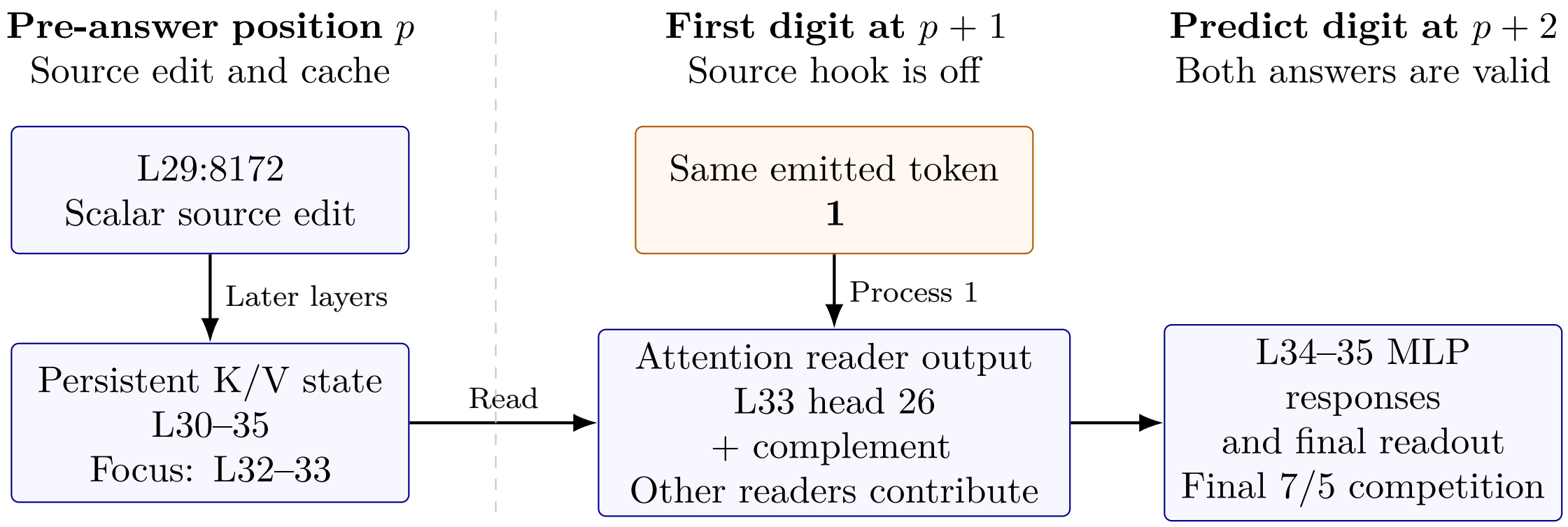
\begin{figure}[tbp]
\centering\begin{tikzpicture}[>=Latex,
 stage/.style={draw=blue!55!black,fill=blue!3,rounded corners=2pt,align=center,font=\small,text width=3.5cm,minimum height=1.2cm},
 heading/.style={align=center,font=\small,text width=4.0cm},
 row/.style={align=left,font=\small,text width=14.0cm}]
\node[heading] at (1.8,1.35) {\textbf{Pre-answer position $p$}\\Source edit and cache};
\node[heading] at (7.7,1.35) {\textbf{First digit at $p+1$}\\Source hook is off};
\node[heading] at (12.7,1.35) {\textbf{Predict digit at $p+2$}\\Both answers are valid};
\node[stage] (edit) at (1.8,0) {L29:8172\\Scalar source edit};
\node[stage,fill=orange!6,draw=orange!70!black] (one) at (7.7,0) {Same emitted token\\\textbf{1}};
\node[stage,minimum height=1.5cm] (cache) at (1.8,-2.2) {Persistent K/V state\\L30--35\\Focus: L32--33};
\node[stage,text width=4.2cm,minimum height=1.5cm] (reader) at (7.7,-2.2) {Attention reader output\\L33 head 26 + complement\\Other readers contribute};
\node[stage,minimum height=1.5cm] (choice) at (12.7,-2.2) {L34--35 MLP responses\\and final readout\\Final 7/5 competition};
\draw[->,thick] (edit)--node[right,font=\scriptsize]{Later layers}(cache);
\draw[->,thick] (one)--node[right,font=\scriptsize]{Process 1}(reader);
\draw[->,thick] (cache)--node[above,font=\scriptsize]{Read}(reader);
\draw[->,thick] (reader)--(choice);
\draw[gray!45,dashed] (4.5,1.7)--(4.5,-3.1);
\node[font=\scriptsize,align=center,text width=14.2cm] at (7.25,-3.65) {Arrows: nested interventions or processing order; no exclusive edges or transition probabilities.};
\node[row] at (7.25,-4.4) {\textbf{Source contrast:} unedited 17; source-edited 15. Reader replacements probe partial effects in hybrid states.};
\node[row] at (7.25,-5.3) {\textbf{One held-out prompt:} the same coordinates retain signed influence; every tested continuation remains 17.};
\end{tikzpicture}
\caption{Partial localization across an unchanged token. The pre-answer edit changes later cached state at position $p$. With the hook removed, both runs process 1 at $p+1$ and predict the next digit. The source contrast is 17 versus 15; reader replacements in the specified hybrid backgrounds can instead restore 17. Arrows summarize nested interventions and processing order, not exclusive edges or transition probabilities. Late-MLP response clamps change relative reader effects without changing their measured answers. Other cache windows and the complement remain influential.}
\label{fig:overview}
\end{figure}

We constructed source-off and source-on prefix caches and verified that both root logits naturally select 1. In this cache panel, transfer inserts source-on states into a source-off recipient; rescue inserts source-off states into a source-on recipient. We then exchanged selected source-position entries before processing this common token and greedily generated the rest of the answer without supplying its second digit. The cache contains state from the pre-answer position, while the later attention query processes the emitted 1 (Figure~\ref{fig:overview}). This panel tests a controlled continuation; the preceding dose scan separately checked full-chat agreement.

Replacing all affected L30--35 source-position K/V entries reproduced the opposite next-token distribution and continuation exactly. Self-swaps and swaps at the unchanged preceding position reproduced the recipient. Full-state transport is an architectural positive control in this fixed-token setting. The selective replacements compare how much each tested cache window changes the continuation margin in the specified recipient.

Among three tested two-layer windows, L32--33 produced the largest displacement: about 68\% of the source margin difference in each direction. L30--31 and L34--35 also contributed. Inserting the edited L32--33 state into the unedited recipient moved the margin from $+0.389778$ to $+0.104454$ nat, leaving 17 as the answer. The reverse replacement moved the edited margin from $-0.029976$ to $+0.257008$ and restored 17. All three partial windows restored 17 in rescue, while none alone transferred the answer 15. Signed margin effects therefore distinguish the strongest tested carrier where answer-flip counts do not. The edited endpoint is only 0.029976 nat from the pairwise zero crossing, whereas the unedited endpoint is 0.389778 nat away. Similar bidirectional movements can therefore restore the answer in one direction without transferring it in the other. The completed continuations verify these outcomes separately from the pairwise margin.

Separate value (V) replacements produced about 54\% of the source movement, compared with about 14\% for keys (K). The V/joint ratio was approximately 0.79--0.80, falling short of the frozen 80\% dominance requirement in at least one direction. At each attention site, keys enter the weighting computation and values the transmitted vector. Across layers, either intervention can also change later queries and processing. These effects localize K/V contributions without separating pure routing from semantic content.

\subsection{Localizing the reader contribution}
Attention reads cached values when processing the shared token. We call its resulting module output the reader output. To follow the larger V contribution, the reader tests used each V-only hybrid with its matching baseline. For the main comparison, the \emph{edited recipient} is the source-on run. Its \emph{V-only hybrid} replaces source-position V entries in L32--33 with source-off values, leaving keys and other cache entries unchanged.

Reader \emph{transfer} copies hybrid outputs into the matching recipient; \emph{rescue} restores recipient outputs inside the hybrid. These names specify replacement directions, not effect signs. In the main direction, transfer moves the 7/5 margin toward 7, reversing part of the original source-induced change toward 5. Restoring both L32--33 reader outputs recovered the matching recipient endpoint; transferring both hybrid outputs recovered the hybrid endpoint. Joint recovery checks the implementation and composition of these interventions: restoring all relevant attention outputs in the defined hybrid backgrounds partly builds recovery into the construction.

The comparisons that split cache windows, keys from values, layers and heads provide the finer localization evidence. Splitting the readers showed a larger L33 contribution, with a remaining contribution from L32. L33 is not the earliest possible reader: saved captures already differ at L30 attention. We next test whether the larger L33 effect concentrates within its head outputs, using the same V-only hybrid backgrounds.

A saved projection proxy nominated query head 26 before we tested its actual 128-dimensional pre-output-projection output slice. We compared the candidate with its complement (the other 31 L33 query heads), head 0 and the all-L33 reference. The projection proxy ranks candidates by an unsigned quantity; the subsequent interventions estimate signed effects. This head-selection procedure is separate from the signed-mean source-neuron nomination.

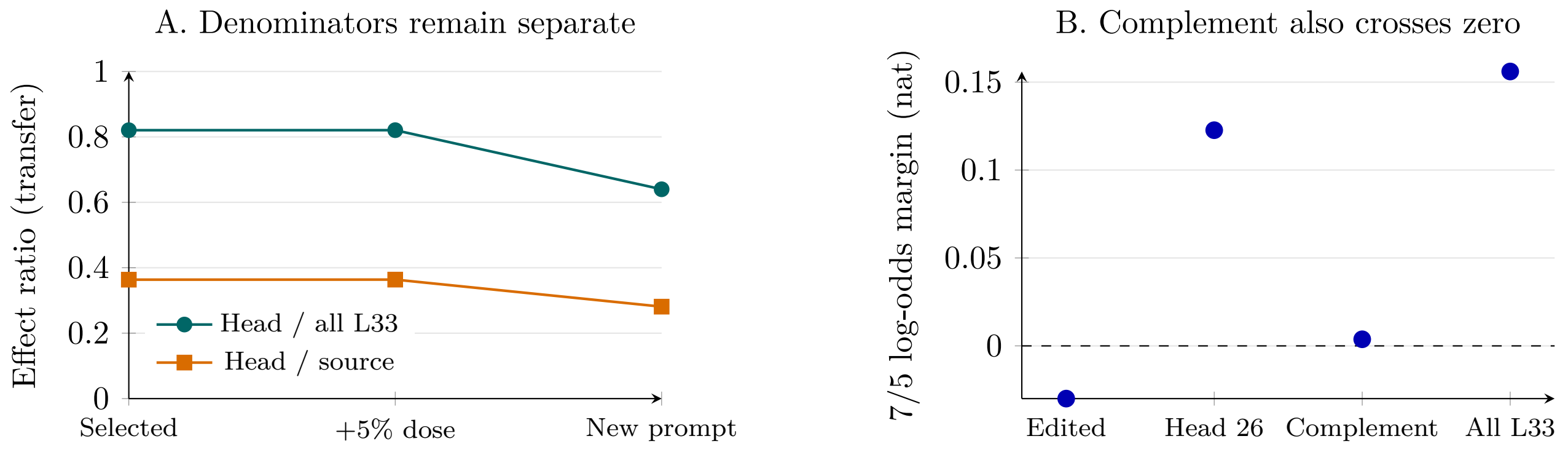
\begin{figure}[tb]
\centering\begin{tikzpicture}
\begin{axis}[width=.43\linewidth,height=5.1cm,axis lines=left,tick label style={font=\small},label style={font=\small},title style={font=\small},legend style={font=\scriptsize,draw=none},ymajorgrids,grid style={gray!20},xtick={1,2,3},xticklabels={Selected,+5\% dose,New prompt},x tick label style={font=\scriptsize},ymin=0,ymax=1,ylabel={Effect ratio (transfer)},title={A. Denominators remain separate},legend pos=south west]
\addplot[teal!80!black,mark=*,thick] coordinates {(1,0.820523337) (2,0.820616345) (3,0.639928656)};
\addlegendentry{Head / all L33}
\addplot[orange!85!black,mark=square*,thick] coordinates {(1,0.36362645) (2,0.363783354) (3,0.281086232)};
\addlegendentry{Head / source}

\end{axis}
\end{tikzpicture}\hfill
\begin{tikzpicture}
\begin{axis}[width=.43\linewidth,height=5.1cm,axis lines=left,tick label style={font=\small},label style={font=\small},title style={font=\small},legend style={font=\scriptsize,draw=none},ymajorgrids,grid style={gray!20},xtick={1,2,3,4},xticklabels={Edited,Head 26,Complement,All L33},x tick label style={font=\scriptsize},ylabel={7/5 log-odds margin (nat)},title={B. Complement also crosses zero},scaled y ticks=false,yticklabel style={/pgf/number format/fixed}]
\addplot[blue!70!black,only marks,mark=*,mark size=2.5pt] coordinates {(1,-0.0299758911) (2,0.122657776) (3,0.0037612915) (4,0.156044006)};
\addplot[black,dashed] coordinates {(0.7,0) (4.3,0)};

\end{axis}
\end{tikzpicture}
\caption{A partial contribution and its boundaries. A: head-26 transfer relative to all L33 readers and to the full source effect, computed separately in each setting. These ratios are not probabilities and are not multiplied along Figure~\ref{fig:overview}. B: replacing the complement can cross the selected 7/5 boundary as well. Points record actual intervention margins; no prompt-population error bars are implied.}
\label{fig:reader}
\end{figure}

Transferring head 26 from the V-only hybrid into the edited recipient moves the next-token margin by $+0.152634$ nat to $+0.122658$, returning the answer to 17. This is about 82\% of the all-L33 effect, but only 36\% of the original source movement. The complement moves the margin to $+0.003761$ and also produces 17 (Figure~\ref{fig:reader}B). Thus a behavioral pivot does not uniquely identify the head with the largest measured contribution. Head-only rescue inside the hybrid also leaves the answer at 17; it does not recover the edited answer 15.

Control head 0 changed the margin by at most 0.0000763 nat, but its displacement was only about 4.4\% of the candidate's. The control therefore tests a smaller perturbation and leaves magnitude-matched specificity unresolved. Other heads in the same KV group changed in the saved geometry and were not all tested individually. The result establishes a concentrated contribution in the nominated head within this panel.

\subsection{Later MLP responses transform reader influence}\label{sec:reader-processing}
Earlier first-digit experiments also found that later MLP computation changes the response to the source across contexts (Appendix~\ref{app:context-response}). Those studies use different intervals and interventions. Here we test the downstream processing of the localized reader effects in the selected continuation setting.
The immediate readout does not explain the difference between head 26 and its complement. Retrospective accounting gives similar norms of the L33 attention-output perturbations and similar immediate contributions to the 7/5 readout, yet head 26's final margin movement is approximately 4.52 times that of the complement (Table~\ref{tab:reader-processing}). This accounting locates a difference after the immediate write but does not isolate its cause. Self-repair research distinguishes normalization changes from later-neuron contributions \cite{selfrepair}, motivating a separate causal test of the responses after the reader. We therefore froze a response-blocking test: while transferring either reader into the edited recipient, we held the L34 and L35 MLP output vectors at their values in that recipient without reader replacement. The clamp acts only while processing the shared first digit. Attention and final normalization remain active, and the baseline MLP contributions are preserved.

\begin{figure}[tb]
\centering\begin{tikzpicture}
\begin{axis}[width=.43\linewidth,height=5.1cm,axis lines=left,tick label style={font=\small},label style={font=\small},title style={font=\small},legend style={font=\scriptsize,draw=none},ymajorgrids,grid style={gray!20},xtick={1,2},xticklabels={Head 26,Complement},xmin=.75,xmax=2.25,ymin=0,ymax=.19,ylabel={Reader effect on margin (nat)},title={A. Selected prompt},legend pos=north east]
\addplot[teal!80!black,mark=*,thick] coordinates {(1,0.152633667) (2,0.0337371826)};
\addlegendentry{Normal}
\addplot[orange!85!black,mark=square*,thick] coordinates {(1,0.0333251953) (2,0.0308456421)};
\addlegendentry{MLP outputs clamped}

\end{axis}
\end{tikzpicture}\hfill
\begin{tikzpicture}
\begin{axis}[width=.43\linewidth,height=5.1cm,axis lines=left,tick label style={font=\small},label style={font=\small},title style={font=\small},legend style={font=\scriptsize,draw=none},ymajorgrids,grid style={gray!20},xtick={1,2},xticklabels={Head 26,Complement},xmin=.75,xmax=2.25,ymin=0,ymax=.19,ylabel={Reader effect on margin (nat)},title={B. Reused wording},legend pos=north east]
\addplot[teal!80!black,mark=*,thick] coordinates {(1,0.101280212) (2,0.0566749573)};
\addlegendentry{Normal}
\addplot[orange!85!black,mark=square*,thick] coordinates {(1,0.0192451477) (2,0.0292434692)};
\addlegendentry{MLP outputs clamped}

\end{axis}
\end{tikzpicture}
\caption{Later MLP responses change the relative influence of the two reader interventions. Points show head-26 and complement effects on the 7/5 margin relative to the matching no-reader recipient; lines connect interventions for comparison, not an ordered computational path. The output clamp holds L34--35 MLP vectors at recipient values. A: the normal head/complement gap nearly disappears. B: the ordering reverses in the reused prompt, although both effects remain positive. Every reader condition shown continues with 17. The experiment comprised twelve core conditions and eight reference/identity conditions across two reused contexts; the figure displays the head and complement effects on the same vertical scale.}
\label{fig:processing}
\end{figure}
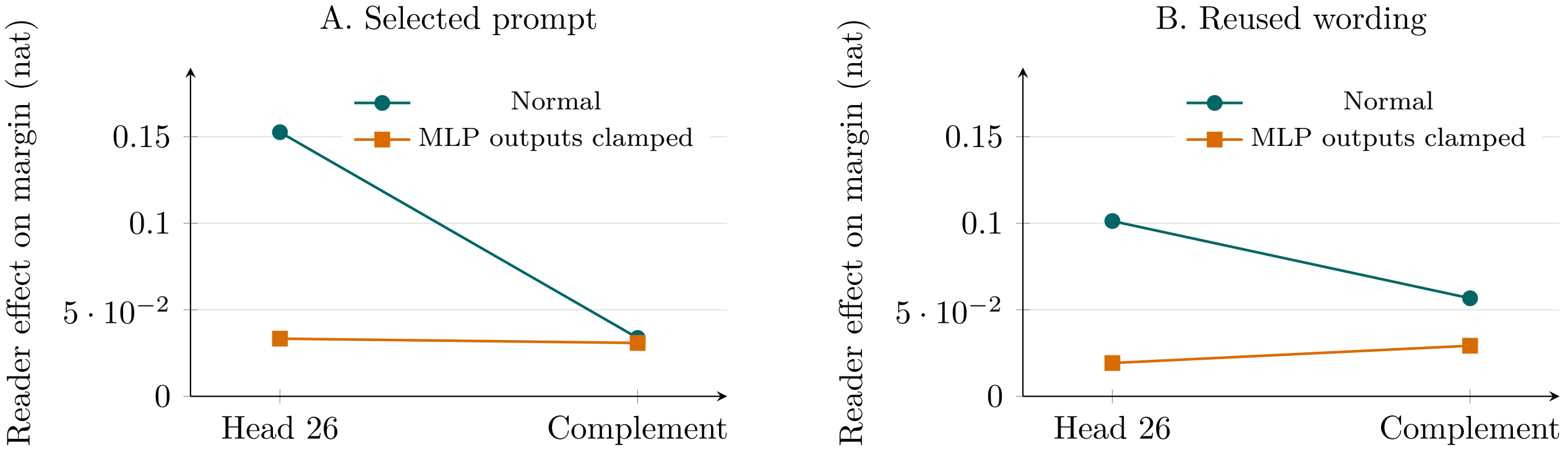

Let $E_H$ and $E_C$ be the head-26 and complement effects on $m$, each relative to the matching no-reader condition. Figure~\ref{fig:processing} compares these effects with normal and clamped processing. Jointly clamping the two MLP outputs reduces their difference $E_H-E_C$ from $+0.118896$ to $+0.002480$ nat in the selected prompt (Figure~\ref{fig:processing}). Thus the selected head's larger final effect depends on the tested late-MLP responses. Both reader interventions still produce 17 with the clamp: their margins are $+0.003349$ and $+0.000870$ nat, just above the pairwise boundary. This measured restoration survives the joint response clamp in the selected condition, with baseline MLP outputs still present. Separate readouts below test the response when only one MLP is held fixed.

The panel also reused the earlier held-out prompt from Section~\ref{sec:route-validation}, with its recorded dose and context-specific donor states. There, clamping changes the head-minus-complement effect from $+0.044605$ to $-0.009998$ nat. This reverses the relative ordering of two positive reader effects; every core condition still produces 17. The second context is an operating-context check, not a fresh holdout. The joint intervention shows that the head/complement effect difference depends on these MLP responses in the tested backgrounds. The retained pre-clamp outputs also allow a post hoc separation of the two MLPs, described next; neither comparison establishes source specificity.

\begin{table}[ht]
\centering\small
\begin{tabular}{lrr}
\toprule
Transfer into edited recipient & Head 26 & Complement\\
\midrule
L33 attention-output change, $\ell_2$ norm & 1.739 & 1.733\\
Immediate readout contribution (nat) & $+0.02237$ & $+0.02322$\\
Final 7/5 margin movement (nat) & $+0.15263$ & $+0.03374$\\
\bottomrule
\end{tabular}
\caption{Similar immediate reader contributions produce different final effects in the selected prompt. Saved-state accounting uses the projected attention-output change and an inferred recipient normalization for the immediate contribution. Final movements are relative to the edited no-reader recipient. Both interventions restore the completed answer 17. Figure~\ref{fig:processing} tests the role of subsequent MLP responses.}
\label{tab:reader-processing}
\end{table}

\paragraph{Separating the two late-MLP responses.}
The saved final residuals and pre-clamp L35 outputs permit separate L34-only and L35-only readouts without another model run (Appendix~\ref{app:single-mlp}). Table~\ref{tab:single-mlp} compares both readers in the selected prompt using common FP64 arithmetic. Either single clamp weakens the head's effect. For the complement, holding L34 strengthens the effect, whereas holding L35 weakens it. The L35-only complement effect remains positive, but is too small to overcome the edited recipient's negative 7/5 margin. Its reconstructed final margin is $-0.019979$ nat; this does not establish a complete answer of 15.

\begin{table}[ht]
\centering\small
\begin{tabular}{lrrrr}
\toprule
MLP output held & Head effect & Head $m$ & Complement effect & Complement $m$\\
\midrule
Neither & $+0.152637$ & $+0.122656$ & $+0.033738$ & $+0.003757$\\
L34 only & $+0.088187$ & $+0.058206$ & $+0.062106$ & $+0.032125$\\
L35 only & $+0.068869$ & $+0.038888$ & $+0.010002$ & $-0.019979$\\
L34 and L35 & $+0.033326$ & $+0.003345$ & $+0.030854$ & $+0.000873$\\
\bottomrule
\end{tabular}

\caption{Separate late-MLP readouts for both readers in the selected prompt, in nats. All entries use the same FP64 reconstruction and corresponding no-reader baseline. Normal and joint-clamp arms also have measured continuations, all 17; the single-clamp arms have only reconstructed readouts. Small differences from the measured FP32 margins elsewhere in the text reflect arithmetic. A negative $m$ means 5 outranks 7, without establishing a full-vocabulary winner or completed answer.}
\label{tab:single-mlp}
\end{table}

Direct-write signs do not determine the effect of clamping a module. The complement's normal L34 and L35 writes have opposite signs in both contexts, but only in the selected context do the two single clamps change its effect in opposite directions relative to normal processing. Appendix~\ref{app:single-mlp} retains the direct-write decomposition and all four reader/context comparisons.

The four clamp conditions expose a further dependence on the computational background. In the reconstructed comparison for the reused context, leaving the L34 MLP output unclamped increases the complement effect by $+0.010646$ nat when the L35 MLP output is unclamped, but decreases it by 0.011835 nat when that output is held at its recipient value. Thus even within one reader/context pair, L34 has no fixed supporting or opposing role under these interventions. The difference between these conditional effects is $+0.022482$ nat. Clamping L34 can change subsequent attention and the input to L35; the contrast does not isolate a neuron-to-neuron edge. It is a comparison of reconstructed readouts, not additional generated answers.

Direct final-residual captures further show that scalar normalization alone does not explain the large head/complement separation under the recipient-divisor accounting (Appendix~\ref{app:reader-accounting}). This decomposition is distinct from an intervention on normalization.

\subsection{Influence persists without an answer change}\label{sec:route-validation}
Before the response-blocking follow-up, we froze a validation panel with the original setting, a 5\% higher dose, and one new prompt at that higher dose: ``Select a single integer from 1 through 30, inclusive.'' The new context had to naturally answer 17, supply an eligible prefix and retain first digit 1 under editing. It qualified, and all thirty conditions completed. Coordinates remained fixed while donor states were recaptured in each context. This design tests reuse of the coordinate and intervention rule with context-specific activations.

In the selected prompt, all-L33 rescue ends at $m=+0.014328$ at the base dose and $m=-0.004700$ at the higher dose, producing 17 and 15 respectively. Head-only rescue retains 17. Nearly unchanged relative contributions can coexist with different answer outcomes near a boundary.

The new prompt starts much farther from that boundary: its source edit moves $m$ from 10.930855 to 10.570538. Head rescue and transfer move it by about $-0.100971$ and $+0.101280$ nat. The transfer contribution is now about 64\% of all L33 and 28\% of the source movement; all continuations remain 17. The signed influence survives, but the panel does not establish a portable answer switch. Wording and natural prefix change together. This validation panel did not test the complement or control head on the new prompt; the subsequent response-blocking panel added the complement comparison reported above, without adding a magnitude-matched control head.

\begin{table}[ht]
\centering\small
\begin{tabular}{lrrrr}
\toprule
Setting & Starting $m$ & Source change & Edited $m$ & Reply\\
\midrule
Selected prompt & $+0.389778$ & $-0.419754$ & $-0.029976$ & $17\to15$\\
Same prompt, +5\% dose & $+0.389778$ & $-0.441185$ & $-0.051407$ & $17\to15$\\
Held-out wording & $+10.930855$ & $-0.360317$ & $+10.570538$ & $17\to17$\\
\bottomrule
\end{tabular}
\caption{Starting competition and measured source movement after the common first digit 1. Margins are 7/5 next-token log odds in nats; completed continuations supply the reply column. The held-out wording uses the higher dose and recaptured donors. These comparisons do not isolate starting margin from the other context changes.}
\label{tab:starting-margin}
\end{table}

The source movement exceeds the starting positive pairwise margin in the two selected-prompt settings, but consumes only 3.3\% of that margin on the held-out wording (Table~\ref{tab:starting-margin}). This arithmetic explains how the recorded signed movement coexists with an unchanged winner. It is not a mediation fraction or a margin-only model of prompt transfer: wording, prefix, dose and donor states also differ.

\paragraph{A fresh check of downstream processing.}
We subsequently froze twelve new wordings with the source, head and base dose $\delta=1.81884765625$ fixed. Eligibility required a completed, format- and token-aligned answer of 17, naturally shared first digit 1 under both source states, and a source write within the 10\% cap. One wording qualified; eight completed with other valid integers and three reached the 80-token cap without a completed choice. The excluded contexts received no intervention. Appendix~\ref{app:fresh-panel} gives the exact rules, all twelve prompts and the complete screening outcomes.

The eligible wording, F10, naturally produced \texttt{I choose **17**.}, with parsed choice 17. Its source edit moved the 7/5 margin from $+5.852978$ to $+5.437187$ nat. The head/complement effect gap was $+0.147114$ nat under normal processing and $+0.010330$ with joint L34--35 clamping, a reduction of $0.136784$ nat (Table~\ref{tab:fresh-reader}). All ten measured continuations remained 17. This adds one conditional recurrence of the later-processing effect, without repeating the full cache-localization sequence or producing another answer switch. The 1/12 count describes coverage of this designed bank's eligibility rule, not a mechanism success rate.

\paragraph{Partial control across generation stages.}
The sequence identifies different sites at which an imposed preference change can be redirected. The source is a scalar MLP activation; the carriers and readers are cached vectors and attention outputs. An unchanged emitted token can therefore conceal a changed continuation state. The complement and held-out tests show that substantial influence can coexist with alternative contributions or an unchanged answer. Together, the interventions explain part of the selected change across positions, while leaving the intervening neuron-level circuitry unresolved. A separate high-dose paired-edit example contrasts earlier state editing with terminal root and suffix editing (Appendix~\ref{app:paired-example}); its coordinates are not additional nodes in this pathway. The next question concerns natural variation: how much of an instruction or history contrast transfers through the same steering target?

\section{How much natural variation transfers through the steering source}\label{sec:natural-transfer}
The preceding experiments impose an activation change and trace part of its effect. The original preference problem also asks how context changes the unedited choice. We therefore replaced the selected activation with values recorded under a different instruction or history and measured movement toward that context's output distribution. This tests whether a useful steering target carries the natural contrast. Che and Li \cite{featureuse} report that strong steering can coexist with weak transfer at donor-observed feature values. We examine this distinction for numerical-choice distributions, with separate diagnostics of activation displacement and local sensitivity.

For donor/recipient pairs, we use progress toward the donor distribution based on Jensen--Shannon (JS) divergence:
\begin{equation}
 F=1-\frac{\sum_j\JS(p_j^{\mathrm{patched}},p_j^{\mathrm{donor}})}
 {\sum_j\JS(p_j^{\mathrm{recipient}},p_j^{\mathrm{donor}})}.
 \label{eq:transfer}
\end{equation}
Negative progress indicates movement away from the donor. This pooled, divergence-based score depends on the scored distribution and is not a linear fraction of log-odds movement. We therefore retain the readout, cohort and matched comparison for each transfer result; Appendix~\ref{app:transfer-diagnostics} illustrates the metric and reports supplementary diagnostics \cite{patching}.

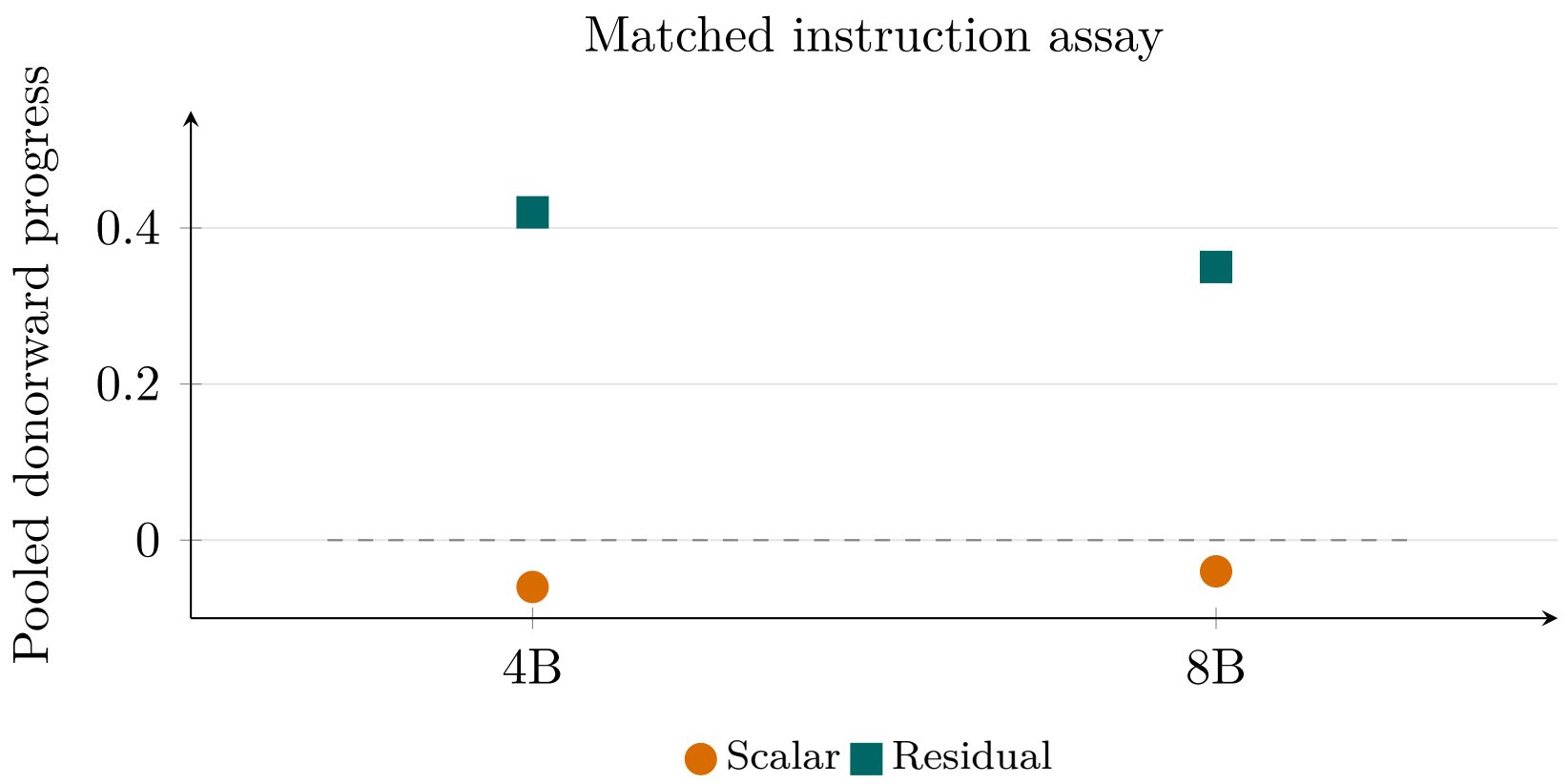
\begin{figure}[tb]
\centering\begin{tikzpicture}
\begin{axis}[width=.65\linewidth,height=5.1cm,axis lines=left,tick label style={font=\small},label style={font=\small},title style={font=\small},legend style={font=\scriptsize,draw=none},ymajorgrids,grid style={gray!20},xtick={1,2},xticklabels={4B,8B},xmin=.5,xmax=2.5,ymin=-.1,ymax=.55,ylabel={Pooled donorward progress},title={Matched instruction assay},legend style={at={(.5,-.2)},anchor=north,legend columns=2}]
\addplot[orange!85!black,only marks,mark=*,mark size=3pt] coordinates {(1,-0.06) (2,-0.04)};
\addlegendentry{Scalar}
\addplot[teal!80!black,only marks,mark=square*,mark size=3pt] coordinates {(1,0.42) (2,0.35)};
\addlegendentry{Residual}
\addplot[gray,dashed] coordinates {(0.7,0) (2.3,0)};

\end{axis}
\end{tikzpicture}
\caption{Scalar and residual replacements on the same ordinary/random pairs, using the same candidate-event readout. The broader residual replacement produces greater donorward progress. This is a matched comparison within the instruction assay; it does not identify a minimal mediator.}
\label{fig:limits}
\end{figure}

A matched instruction sweep compared scalar and broader-state replacements on ten ordinary-versus-random templates at 1--30, yielding twenty directed pairs. L29 residual replacement gave candidate-event progress of 0.42 and 0.35 for 4B and 8B, compared with $-0.06$ and $-0.04$ for scalar swaps (Figure~\ref{fig:limits}). The broader-vector replacement produced more donorward progress in this matched assay. Its different cohort and instruction wording keep it separate from the ordinary/uniform bank; identifying a minimal mediator would require further interventions.

In the ordinary-versus-uniform confirmation bank, pooled candidate-event progress was 0.009, 0.017 and 0.001 for the three checkpoints, against natural total variation (TV) contrasts of 0.66, 0.65 and 0.71. Median swap TV was 0.011, 0.013 and 0.016. The small pooled effects included substantial tails: 7--13\% of swaps moved more than 0.1 TV, with both donorward and opposing movements. The distribution of effects therefore matters alongside its average.

History tests replaced decision-position activations after different earlier values. Across 32 directed swaps, pooled progress was 0.0007 [0,0.0016] in 4B and 0.0014 [0,0.0034] in 8B, using a prompt bootstrap. Broader cache tests used sixteen directed pairs over eight prompts for earlier values 17 and 23, with the next leading digit fixed to 1 or 2. Replacing earlier-number K/V states across layers gave conditional continuation progress of 0.753 and 0.903. Replacing these earlier-number cache states transferred substantial donorward variation in the fixed-leading-digit continuation assay. Its cohort and conditional endpoint differ from the scalar summary, so the fractions do not form an additive decomposition.

Post hoc displacement and sensitivity diagnostics help describe the weak average transfer, with substantial individual effects and cancellation. Appendix~\ref{app:transfer-diagnostics} reports those readout-specific analyses separately from the donorward progress above.

For the tested instruction and history contrasts, swapping the decision-position activation transferred little of the output change on average. This result coexists with strong steering effects in other settings and substantial effects in some individual swaps. Earlier positions, multiple components and context-dependent processing remain possible contributors to the natural contrasts.

Changing a valid answer does not by itself meet a requested sampling law. Appendix~\ref{app:distribution} retains separate uniformity and history-dependence analyses, including limits of a controller restricted to leading-digit weights. Those results concern different assays and do not define the selected continuation pathway.

\Needspace{10\baselineskip}
\section{Discussion}
Numerical choice makes preference measurable while holding answer validity explicit. The equal-validity design isolates the outcome being studied; it does not establish machinery exclusive to preference. The fixed-dose panel tests directional expectations, and retrospective response profiles estimate much of the probability redistribution. Their selected prediction miss shows why distributional accuracy and winner prediction must be assessed separately: a familiar response can cross a narrow boundary between two plausible answers. The held-out source effect leaves a much larger positive margin and the answer unchanged. That comparison does not isolate starting margin from the accompanying wording, prefix and donor changes.

The unchanged first digit provides a controlled view of how an earlier edit affects later choice. The source hook has been removed when the model processes 1, yet the altered cached state changes the competition between 7 and 5. Full-cache replacement checks state transport; selective window and reader replacements localize contributions. Similar immediate reader writes then receive different downstream treatment. Joint late-MLP clamping changes their effect gap while the measured replies remain 17, and reconstructed single-MLP contrasts show that a component's signed effect depends on the later computational background. These findings explain part of an imposed preference change across positions without establishing exclusive edges or a symbolic representation of the intended complete number.

Natural-context replacements bound the explanatory reach of the steering source. Small average scalar transfer coexists with substantial individual effects and with greater progress from broader-state replacements. The selected activation can influence a choice while carrying little of the measured instruction or history contrast by itself. Other components and context-dependent processing remain possible contributors. The partial pathway, active complement and limited prompt coverage support a conditional account of preference processing.

\paragraph{Relation to prior work.}
Numerical-sampling studies distinguish independent requests from history-conditioned generation and report failures to match requested laws \cite{numbers,stochastic,renda,dice}. Renda et al.\ \cite{renda} already score candidate probabilities and measure valid-domain containment. Base/aligned comparisons address changes in numerical diversity \cite{ansari,west}, while response-history studies show that marginal bias can change across turns \cite{bscore}. Tool and programming-language studies motivate the broader question of choice beyond correctness \cite{tools,languages}. Our task makes validity explicit while permitting a causal analysis of probability allocation; it does not establish that the numerical mechanism transfers to those applications.

Numerical mechanisms also have direct precedents. Nikankin et al.\ \cite{heuristics} identify arithmetic neurons with numerical-pattern heuristics. Shao et al.\ \cite{benford} identify late digit-selective FFN neurons and use targeted pruning to change digit bias in numerical reasoning. Levy and Geva \cite{base10} provide causal evidence for digit-based representations, and Baeumel et al.\ \cite{modular} intervene on digit-position circuits in arithmetic. Arora et al.\ \cite{sparse} demonstrate single-neuron dose-dependent answer changes in their Section 6.2 and analyze arithmetic digit features in Appendix H.1. We connect the induced answer change to selective cache and attention-output interventions after an unchanged emitted token, then test how later MLP responses change the readers' relative influence. This continuation-level account uses established discovery and steering methods; the combined nomination procedure is not evaluated for superiority.

Steering-strength analysis and self-repair research explain why an edit's final effect can differ from its immediate write \cite{hydra,strength,selfrepair}. The head/complement comparison tests a specific instance: late-MLP responses change their relative margin effects while the corresponding generated answers remain fixed under the joint clamp. Activation-patching and natural-value studies further distinguish effective manipulation from explanation of natural behavior \cite{vig,patching,illusion,featureuse}. The natural-context swaps bound how much unedited variation the steering source carries. Together, these comparisons support a conditional account of the selected pathway, with limited prompt replication and no claim to identify a minimal faithful circuit.

\paragraph{Scope and limitations.}
The supporting response studies cover three related Qwen checkpoints; the detailed source/cache/reader pathway uses Qwen3-4B. The prompts are English requests over small intervals. Candidate scores depend on the response interface and admit a nondigit as a terminator; generated-answer checks cover their stated conditions. An earlier six-question serialization comparison illustrates the size of an interface effect (Appendix~\ref{app:interface}). The second-digit switch prompt was selected near a 17/15 boundary, and the held-out route test used one eligible prompt with context-specific donor activations. Separate late-MLP readouts were reconstructed after the frozen joint-clamp experiment; they do not supply additional generated-answer observations. Source, cache and head fractions use different denominators and cannot be multiplied into a path probability. Head controls leave magnitude-matched specificity unresolved, and complementary components remain influential. Edited hybrid states are counterfactual constructions whose natural occurrence remains untested. The corrected nomination verifies the 4B coordinate; the other coordinates retain their historical magnitude-based selection. The full records behind the earlier interval predictor remain unavailable, and the retrospective shared-profile comparison is not a replacement validation.

The fresh screen in Section~\ref{sec:path} adds one conditional recurrence but leaves eleven contexts without intervention evidence. Requiring an unedited answer of 17 restricts coverage more than the 7/5 readout itself requires, including by excluding five completed 15 replies. With twelve designed prompts and one greedy baseline each, the counts are descriptive; no population recurrence rate is inferred. Wording, generated preambles and baseline competition may reflect deeper context-dependent processing, but this panel does not isolate their contributions.

The origin of the natural 17 preference and the upstream inputs that recruit the source activation remain open. Establishing an exclusive gate or a minimal circuit would require comparisons beyond this panel. Additional model families, all-number steering, adaptation and distribution-preserving masks address further questions. The present evidence supports a focused account of how an imposed neuron edit changes preference through downstream computation and token competition. Transfer to other tasks with multiple valid answers remains a question for future work.

\Needspace{8\baselineskip}
\paragraph{Conclusion.}
Numerical choice provides a controlled setting for studying preference among valid alternatives. A selected pre-answer edit changes 17 to 15 across an unchanged first token; selective cache and reader replacements localize part of its influence. Later MLP responses change the relative reader effects even when the measured answers remain fixed. The probability-response and prompt tests distinguish structured influence from a portable answer switch. Together with natural-context transfer, they support a conditional account of how an imposed preference change is carried and processed during inference. The origin and complete circuit of unedited numerical preference remain unresolved.

\clearpage\appendix
\section{Experimental protocols, cohorts and evidence status}\label{app:cohorts}

\subsection{Selection and evaluation cohorts}
In the 4B nomination procedure, scores are normalized by each branch's all-layer maximum within each range edit and summed over edits. The union of the top twenty coordinates from each branch is reduced to twenty by the sum of normalized scores. The zeroing screen selects among eligible coordinates by the increase of $q(17)$ at 1--40; layer and neuron index break ties. The protected-range rule and the corrected coordinate's branch ranks are given in Section 2. Earlier exploratory ablations already showed the selected unit's 1--40 effect. The frozen screen formalized comparison with alternatives, while later new-wording and interval tests supplied separate evaluation. Selection ranges, new wording and new intervals remain separate evaluation categories.

The earlier zeroing study distinguishes new paraphrases from eight templates previously used in exploration. Eligible reused counts are 8, 7 and 8. Their sampled natural/zeroed 17 fractions are 0.25/0.57, 0.00/0.42 and 0.32/0.75. For free generation on the new cohort, using eight replies per template, the natural/zeroed fractions are 0.04/0.90, 0.00/0.76 and 0.17/0.92. The intervention acts at every position in the free-generation assay. The two cohorts are reported separately. Eligibility required a naturally generated bare or emphasized integer at the answer boundary; the FP32 confirmation accepted a final period or exclamation mark without adding observations. The combined new-plus-reused 1--40 cohorts excluded 4, 4 and 0 of 28 prompts, leaving 24, 24 and 28. These totals do not replace the separate new-wording denominators. The documented intervals use a 95\% template bootstrap; the original resample count and random seed are not available in the retained historical summaries.

The later eighteen-prompt scalar comparison includes ninety source/control/self recipient cells, with five arms per prompt. These are not ninety independent prompts. The frozen 216-condition panel includes two targets, two matched controls, self and natural-donor arms. All 72 fixed target doses, 72 control doses and 36 self conditions were measured; seventeen of 36 natural patches exceeded the write cap. E6's range-35 donor was truncated before its answer and automatically mislabeled from a repeated bound. Both donor patches were already skipped. Its recipient fixed-dose measurements remain valid; excluding the entire pair leaves 66 predicted effects and two unresolved among 68 target doses. The primary 17/23 signs were internally registered; the later 17/15 analysis and switch selection were exploratory. The downstream collection consists of separate frozen packets: dose scan 18 conditions, cache window 14, K/V split 12, joint reader 12, reader split 20, head panel 24, held-out validation 30 and late-MLP response blocking 20 (twelve core cells plus eight reference/identity conditions). That response-blocking panel reuses two already studied contexts and is not a new prompt holdout. The subsequent twelve-wording screen is reported separately in Appendix~\ref{app:fresh-panel}. Repeated baselines and controls across packets are not independent replications.

The scalar instruction bank uses ordinary/equal-probability requests; the matched residual sweep uses ordinary/random requests in ten templates. The history scalar summary includes two value-pair contrasts, whereas the cache assay uses 17 versus 23, fixed leading digits and sixteen directed pairs. These cohorts do not isolate contrast type as the only changed variable.

\subsection{Frozen directional-panel specification}\label{app:fixed-protocol}
The directional panel used Qwen3-4B in the pinned streamed-FP32 runtime described in Appendix~\ref{app:evidence}. L29:8172 was the gradient-branch nominee and zeroing-screen winner. L34:2836 was a retained historical candidate whose natural range-35 patch effects had opposing signs across wording groups and nearly canceled when pooled. That observation, together with negative local gradients in all twelve historical contexts, motivated the separate fixed-dose comparison. The target selection used earlier outcomes; neither target was nominated on the six extension responses.

\begin{table}[ht]
\centering\small\begin{tabular}{lrlr}
\toprule
Target & Positive dose & Control & Positive control dose\\
\midrule
L29:8172 & 1.81884765625 & L29:8369 & 1.897157100217\\
L34:2836 & 1.81298828125 & L34:1261 & 2.305332032952\\
\bottomrule
\end{tabular}

\caption{Fixed target and residual-write-matched control doses. Each site was tested individually with both the displayed positive dose and its exact negative. Control magnitudes are rounded to twelve decimal places; the archive retains full precision.}
\label{tab:fixed-protocol}
\end{table}

Each target-dose magnitude is one quarter of the median absolute historical range-35 donor displacement. A seeded draw (Python \texttt{random.Random(172903)}) selected one same-layer control per target, excluding the historical shortlist. The draw did not use control outcomes. Its dose was rescaled once as $\delta_c=\delta_t\lVert W_{\mathrm{down},t}\rVert_2/\lVert W_{\mathrm{down},c}\rVert_2$ to match the residual-write norm. This is a single-coordinate comparison per target, not a population specificity test.

For $m_{17,23}=\log w(17)-\log w(23)$, the frozen prediction was $\Delta m_{17,23}>0$ for a positive L29 dose and $\Delta m_{17,23}<0$ for a positive L34 dose, with reversed directions for negative doses. Only effects with magnitude greater than 0.05 nat counted as resolved. The six arms per target/context were self replacement, positive and negative target doses, a natural donor displacement, and positive and negative control doses. The write cap was 10\% of the recipient whole-MLP output norm; over-cap cells were skipped without changing their doses. Identity tolerances were TV $\leq10^{-4}$ and margin error $\leq0.001$ nat; dose and scoring-consistency tolerances were 0.001 in activation units and nats, respectively.

\begin{table}[ht]
\centering\small\begin{tabularx}{\linewidth}{l>{\raggedright\arraybackslash}X}
\toprule
Context & Exact recipient prompt\\
\midrule
P1 & Give me a number between 1 and 30\\
P2 & Give me a number between 1 and 30.\\
P3 & Give me a number from 1 to 30\\
P4 & Give me a number from 1 to 30.\\
P5 & Pick a number between 1 and 30\\
P6 & Pick a number between 1 and 30.\\
P7 & Pick a number from 1 to 30\\
P8 & Pick a number from 1 to 30.\\
P9 & Choose a number between 1 and 30\\
P10 & Choose a number between 1 and 30.\\
P11 & Choose a number from 1 to 30\\
P12 & Choose a number from 1 to 30.\\
\midrule
E1 & Give me a whole number from 1 through 30.\\
E2 & Give me one integer in the inclusive interval [1, 30].\\
E3 & Pick a whole number from 1 through 30.\\
E4 & Pick one integer in the inclusive interval [1, 30].\\
E5 & Choose a whole number from 1 through 30.\\
E6 & Choose one integer in the inclusive interval [1, 30].\\
\bottomrule
\end{tabularx}

\caption{The eighteen recipient prompts in the fixed-dose panel. P1--P12 retain the historical pre-answer prefixes; E1--E6 use newly generated prefixes. Range-35 donor prompts replace the upper endpoint 30 by 35. Punctuation is part of the prompt.}
\label{tab:fixed-prompts}
\end{table}

Each scored distribution was conditioned on its recorded pre-answer prefix. Historical prefixes were retained from the earlier runs; extension prefixes were generated greedily with the recorded non-thinking chat template and a 48-token cap. The directional panel did not require a baseline answer of 17 or a donor answer of 23. The \href{https://github.com/phunterlau/why17/blob/main/results/fixed_dose_protocol_20261030/protocol.json}{frozen specification}, dated 30 October 2026 at 05:08 UTC, contains the exact prompts, token/prefix records for the historical contexts, endpoints and signed predictions. The \href{https://github.com/phunterlau/why17/blob/main/results/fixed_dose_protocol_20261030/input_provenance.json}{selection and input record} and \href{https://github.com/phunterlau/why17/blob/main/results/fixed_dose_results_20261030/raw/write_calibration.json}{write calibration} specify the control rule and scale. The later 17/15 analysis and switching-case selection were exploratory.
\FloatBarrier

\subsection{Qualification and control summary}\label{app:qualifications}
\begin{table}[ht]
\centering\footnotesize
\begin{tabularx}{\linewidth}{>{\raggedright\arraybackslash}p{2.5cm}XX}
\toprule
Test & Recorded boundary or control & Consequence for the claim\\
\midrule
Original 4B validation & Protected 1--20 answer changes from 13 to 5 & No all-criteria validation success; keep substantial 1--40 effects separate.\\
Original 2507 validation & $q(17)$ increases by 0.28, 0.03 and 0.00 at 1--35, 1--38 and 1--45 & Extension-interval reach criterion fails.\\
Original 8B validation & Some extension baselines already favor 17 & The assumed baseline condition does not apply there.\\
First-digit diagnostic & FP32 pooled share 0.698 versus a 0.70 criterion (bf16: 0.702) & Threshold fails in FP32; distinct from fitted-predictor accuracy.\\
Zeroing specificity & 100 controls matched on down-projection-column norm & Does not match activation-dependent removed writes.\\
Natural instruction swaps & 16 matched control neurons give comparably small effects & Weak transfer is not unique to the selected source.\\
Fixed-dose source & One residual-write-matched coordinate, with an active opposite effect & Specificity is assessed against this one control.\\
K/V split & V exceeds K; V/joint does not meet 80\% in both directions & Keys retain a contribution alongside values.\\
Head decomposition & Complement can also switch the answer; head-0 displacement is only 4.4\% of the candidate's & Neither a unique gate nor a magnitude-matched head-specificity test.\\
Held-out route & One eligible new prompt retains signed effects; every answer remains 17 & Influence transfers under this assay; answer control does not.\\
Fresh wording screen & One eligible context among twelve attempted wordings; eleven untested by intervention & One conditional recurrence; coverage does not estimate mechanism prevalence.\\
Late-MLP blocking & Joint clamp preserves replies; separate readouts reveal background-dependent effects & Selected prompt: L34-only and L35-only clamps change the complement effect in opposite directions. Reused prompt: the conditional L34 effect reverses with the L35 background. No single-clamp continuations.\\
\bottomrule
\end{tabularx}
\caption{Qualifications belong to their own cohorts and endpoints. The table combines documented historical criteria with later frozen follow-up checks; it does not pool them into a success rate.}
\label{tab:qualifications}
\end{table}
\FloatBarrier

\subsection{Response-interface sensitivity}\label{app:interface}
Holding six Qwen3-4B questions at 1--30 and the prefix ``The number is '' fixed, an earlier serialization comparison gave mean $q(17)=0.284$ with chat markup and 0.854 with an added empty thinking-mode block. No reasoning was generated in that block. This interface effect concerns conditional scores, not the training origin of preference or mediation by the selected pathway. Its cohort differs from the own-prefix baseline and continuation study.

\subsection{Continuous arithmetic controls}\label{app:arithmetic}
An earlier study scored 571 exact-answer problems whose correct values lie in 1--30: addition, subtraction, linear equations and inclusive counting. It used FP32 Qwen3-4B and 8B with a bare short-answer interface and zeroed the selected source. The retained historical summary describes scoring at the ``final answer position'' but does not identify the intervention hook position precisely; we cannot equate it with the later pre-answer-only edit. Table~\ref{tab:arithmetic} retains its documented continuous measurements. The original per-item records were not recovered for the present saved-data reanalysis, so these values remain historical aggregate summaries.

\begin{table}[ht]
\centering\small
\begin{tabular}{lrr}
\toprule
Zeroing measurement & Qwen3-4B & Qwen3-8B\\
\midrule
Mean absolute change in $P(\mathrm{correct})$ & 0.0074 & 0.0044\\
Mean answer-distribution TV & 0.019 & 0.006\\
Items with $|\Delta P(\mathrm{correct})|>0.05$ & 3.4\% & 1.3\%\\
Largest absolute $P(\mathrm{correct})$ change & 0.62 & 0.50\\
\bottomrule
\end{tabular}
\caption{Historical computed-answer controls. Small average movement coexists with a sensitive tail. The task, prefix and zeroing intervention differ from the selected additive-dose continuation experiment.}
\label{tab:arithmetic}
\end{table}

The post hoc tail review found many sensitive 4B cases among near-tie equations, with comparable matched-control effects. These observations do not establish preference specificity: confidence, interfaces and interventions differ from the free-choice study. Supplied-result and thinking-trace extensions are omitted because their answer-location rules carried further qualifications.

\subsection{Implementation and runtime coverage}\label{app:evidence}
Historical discovery used RMS magnitude aggregation, including for 2507 and 8B. The corrected 4B nomination uses absolute signed-mean PP scores with the original gradient aggregation; a sensitivity analysis changes both branches. Both retain L29:8172 through the gradient shortlist. Only 4B was renominated; the other coordinates retain their original provenance. The pathway study subsequently used the known 4B source.

Historical search used bfloat16 weights with an FP32 output head. Confirmation of the principal validation, prediction and interval-response findings used FP32. The pathway study used unquantized FP32 and streamed CUDA: weights reside in host memory and decoder blocks move through the GPU during computation. All experiments ran on an RTX 2070 Max-Q, as reported by the authors; the pathway archives additionally retain device records. Host memory is therefore part of the execution requirements alongside the 8GB-class GPU.

The earlier FP32 confirmation records revision prefixes \texttt{1cfa9a72}, \texttt{cdbee75f} and \texttt{b968826d}. Some earlier discovery runs lacked repository revision records. The later 4B route pins revision \nolinkurl{1cfa9a7208912126459214e8b04321603b3df60c}, Torch 2.11.0+cu128 and Transformers 5.20.1. These later settings must not be retroactively assigned to every earlier experiment.

A separate FP16 batching qualification failed at TV 0.002458 against a 0.001 threshold. It neither replaces the original FP32 confirmation nor becomes qualified by that confirmation.

\subsection{Study records}
Table~\ref{tab:study-index} maps study families to the paper. The \href{https://github.com/phunterlau/why17/blob/main/paper_clean/STUDY_INDEX.md}{repository study index} links protocols, source files, runtime records and hashes. Historical aggregate summaries retain their limited coverage; later continuation results were audited against saved distributions, replies and interventions. Unavailable per-prompt records are not reconstructed.

\begin{table}[ht]
\centering\small
\begin{tabularx}{\linewidth}{>{\raggedright\arraybackslash}p{3.5cm}>{\raggedright\arraybackslash}Xl}
\toprule
Study & Measurements and records & Presentation \\
\midrule
Preference and interface & Conditional baseline scores and fixed-question serialization comparison & Section 2; App.~\ref{app:interface} \\
Discovery and selection & PP/gradient scores, causal eligibility and corrected 4B ranks & Section 2 \\
Wording and interval tests & Historical cohorts, sampling settings and aggregate predictions & App.~\ref{app:historical-prediction} \\
Fixed-dose direction & Two targets, matched controls and frozen 17/23 predictions & Section~\ref{sec:fixed-dose} \\
Branch and continuation & Saved-event KL, retrospective predictions and paired event masses & Fig.~\ref{fig:response-geometry}; Apps.~\ref{app:branch-analysis}--\ref{app:profile-prediction} \\
Computed-answer controls & Historical continuous scores and sensitive tails & App.~\ref{app:arithmetic} \\
Composed edits & Earlier state and terminal readout programs in two contexts & App.~\ref{app:paired-example} \\
Contextual response & Direct/full responses, blocking and distinct cancellation profiles & Figure~\ref{fig:response-detail} \\
Source dose and rivals & Selected prompt, event scores and completed replies & Figure~\ref{fig:dose} \\
Cache and reader localization & Source position, window/K/V swaps, reader and head controls & Figures~\ref{fig:overview}, \ref{fig:reader} \\
Reader processing & Immediate-effect accounting; late-MLP clamps and direct norm captures & Fig.~\ref{fig:processing}; Table~\ref{tab:single-mlp}; App.~\ref{app:single-mlp} \\
First-digit accounting & Historical module and unit responses to source zeroing & App.~\ref{app:first-digit-accounting} \\
Held-out route & Fixed coordinates, higher dose and recaptured donors & Table~\ref{tab:starting-margin} \\
Fresh wording screen & Twelve baselines; one eligible reader/MLP recurrence & App.~\ref{app:fresh-panel} \\
Natural-context transfer & Paired cohorts, scalar/broader-state swaps and diagnostics & Figures~\ref{fig:limits} and~\ref{fig:weak-transfer} \\
Sampling laws & Uniformity, branch-control bounds and second-draw summaries & Figure~\ref{fig:sampling-limits} \\
\bottomrule
\end{tabularx}
\caption{Named study families connect the main results to their experimental records. Appendix~\ref{app:cohorts} gives cohort and control boundaries; each study retains its own numerical profile and endpoint.}
\label{tab:study-index}
\end{table}

\FloatBarrier
\section{Continuation-pathway validation and downstream processing}\label{app:path-support}
\subsection{Retrospective readout accounting}\label{app:reader-accounting}
\paragraph{Earlier inferred-divisor accounting.}
Table~\ref{tab:reader-processing} uses saved attention and MLP outputs at the first-digit query. Let $x_r$ and $x_c$ denote the final residuals in the recipient and intervened conditions, and let $a$ be the difference between the 7 and 5 unembedding rows multiplied elementwise by the final normalization gain. With RMS divisor $\rho(x)=\sqrt{\lVert x\rVert^2/d+\epsilon}$, where $d$ is the residual width and $\epsilon$ the normalization constant, the margin is $m=a^\top x/\rho(x)$. Its change satisfies
\begin{equation}
 m_c-m_r=\frac{a^\top(x_c-x_r)}{\rho_r}
 -\left(\frac{\rho_c}{\rho_r}-1\right)m_c.
 \label{eq:readout-accounting}
\end{equation}
The first term can be split over the captured residual increments. The table's immediate contribution uses only the changed L33 attention output; subsequent attention and MLP updates, together with the second term, contribute to the final movement.

The earlier analysis jointly inferred $1/\rho_r$ and $\rho_c/\rho_r-1$ by least squares across ten centered digit log probabilities and captured residual changes; it did not record each final divisor. Maximum margin-closure errors were $8.52\times10^{-6}$ nat for principal contrasts and $1.13\times10^{-5}$ in the wider head panel. Fitting these divisors to the same outputs makes closure an internal accounting check, not independent causal validation or an isolated normalization intervention.

\paragraph{Direct-capture validation and response blocking.}
Direct final-residual and RMSNorm captures in the response-blocking test also clarify the readout. In the selected prompt, normal head transfer contributes $+0.152607$ nat through the numerator term at the recipient divisor and $+0.000030$ nat through the divisor adjustment. The complement contributes $+0.033741$ and $-0.000002$ nat, respectively. Using the unrounded FP64 terms, the head-minus-complement gap is $+0.118867$ nat from the numerator term plus $+0.000032$ nat from the divisor adjustment, totaling $+0.118899$ nat. The final scalar divisor adjustment therefore does not explain the large separation in this comparison. Earlier normalization can still affect the residual. These terms are descriptive accounting, not separate interventions on RMSNorm.

The subsequent response-blocking test in Figure~\ref{fig:processing} directly captured the final residual, normalization gain, epsilon, normalized hidden state and full next-token distribution. FP32 reconstruction reproduced the captured normalized output bitwise, and readout accounting closed within $1.11\times10^{-5}$ nat. Nine archived endpoint comparisons matched exactly in full next-token distribution, margin and continued token IDs; six identity and six clamp-scope checks passed. All twenty conditions and 186 journaled forwards completed, with no missing or truncated replies. The joint output clamp tests late-MLP responses causally; the numerator/divisor decomposition remains descriptive.

\subsection{Separate late-MLP counterfactual readouts}\label{app:single-mlp}
The final decoder MLP is followed by RMSNorm and the output head, with no further attention or MLP block. Let $x_N$ be the normal final residual, $x_J$ the residual with both MLP34 and MLP35 outputs clamped, and $M_{35,r}$ the no-reader recipient's MLP35 output. Let $M_{35,N}$ be the normal output and $M_{35,J}^{\mathrm{pre}}$ the output captured before replacement in the joint-clamp run. The single-arm residuals can then be reconstructed as
\[
 x_{35}=x_N-M_{35,N}+M_{35,r},\qquad
 x_{34}=x_J-M_{35,r}+M_{35,J}^{\mathrm{pre}}.
\]
The second expression uses an MLP35 output already computed with L34 clamped. For each reconstructed residual, we recompute its RMS divisor using the captured epsilon and apply the captured normalization gain and 7-minus-5 readout. This recovers the corresponding mathematical readout up to saved floating-point roundoff, without a new transformer forward pass or continuation.

Independent CPU reconstruction checked twelve input-tensor hashes and the capture/replacement order. Normal and joint-clamp margins matched measured logits within $1.11\times10^{-5}$ nat; FP32 arithmetic changed the FP64 single-arm results by at most $5.7\times10^{-7}$ nat, preserving the negative final 7/5 margin for the complement reader in the selected prompt when only L35 is clamped. These are arithmetic checks, not uncertainty intervals. The measured joint-clamp margin $+0.000870$ is below the 0.001-nat replay tolerance, but that tolerance is not an error bar: endpoint replays required identical reply tokens and matched exactly.

\begin{table}[ht]
\centering\small
\begin{tabular}{llrrrr}
\toprule
Context & Reader & Normal & L34 held & L35 held & Both held\\
\midrule
Selected & Head 26 & $+0.152637$ & $+0.088187$ & $+0.068869$ & $+0.033326$\\
Selected & Complement & $+0.033738$ & $+0.062106$ & $+0.010002$ & $+0.030854$\\
Reused & Head 26 & $+0.101275$ & $+0.072788$ & $+0.061985$ & $+0.019258$\\
Reused & Complement & $+0.056680$ & $+0.046034$ & $+0.017404$ & $+0.029239$\\
\bottomrule
\end{tabular}

\caption{All four reader/context comparisons under common FP64 reconstruction. Entries are incremental 7/5 effects in nats relative to the matching no-reader background. The two contexts were reused; separate-clamp columns have no generated replies. Opposing single-clamp changes relative to normal occur only for the selected complement.}
\label{tab:mlp-all-cells}
\end{table}

At the recipient normalization divisor, the selected head's L34 and L35 MLP changes contribute $+0.02586$ and $+0.08375$ nat; the complement contributes $-0.01325$ and $+0.02375$. These differences account algebraically for 83.4\% of the normal head/complement gap, not an isolated mediation fraction. Holding L34 also changes subsequent processing: the complement's L35 MLP contribution becomes $+0.03127$, and its L35 attention contribution changes from $-0.00009$ to $+0.00752$ nat.

Let $E_N,E_{34},E_{35},E_J$ denote incremental reader effects under normal, L34-held, L35-held and joint-held processing. Allowing L34 to respond has effects $E_N-E_{34}$ when L35 is free and $E_{35}-E_J$ when L35 is held. Their interaction contrast is
\[
 I=E_N-E_{34}-E_{35}+E_J.
\]
\begin{table}[ht]
\centering\small\begin{tabular}{llrrr}
\toprule
Context & Reader & L35 free & L35 held & Interaction\\
\midrule
Selected & Head 26 & $+0.064449$ & $+0.035542$ & $+0.028907$\\
Selected & Complement & $-0.028367$ & $-0.020852$ & $-0.007515$\\
Reused & Head 26 & $+0.028487$ & $+0.042727$ & $-0.014240$\\
Reused & Complement & $+0.010646$ & $-0.011835$ & $+0.022482$\\
\bottomrule
\end{tabular}

\caption{Conditional L34 effects derived from the same reconstructed readouts as Table~\ref{tab:mlp-all-cells}, in nats. The reused complement changes sign with the L35 background. These are defined output-clamp contrasts, not additive natural-mediation fractions or generated replies.}
\label{tab:mlp-conditional}
\end{table}

The reconstructed head-minus-complement gaps for normal, L34-only, L35-only and joint processing are 0.118899, 0.026082, 0.058867 and 0.002473 nat in the selected prompt, versus 0.044595, 0.026754, 0.044581 and $-0.009981$ in the reused wording. Holding L35 barely changes the latter gap because both effects decrease similarly. An unchanged gap need not imply an inactive component. These values use common FP64 reconstruction; the main figure uses measured FP32 margins.

\subsection{Fresh wording screen for reader/MLP recurrence}\label{app:fresh-panel}
The fresh panel fixed L29:8172, dose 1.81884765625, L33 head 26 and its 31-head complement before collecting the twelve wordings in Table~\ref{tab:fresh-screen}. It used the same pinned Qwen3-4B streamed-FP32 runtime. Each wording received one greedy baseline with thinking disabled and an 80-token cap. The frozen parser had to identify a completed answer of 17, pass its answer-format check and locate a token-aligned numerical boundary. The source-off and source-on states then both had to select first digit 1 naturally, and the source write had to remain within 10\% of the recipient whole-MLP output norm. Failed contexts were retained without replacement, retuning or intervention collection.

\begin{table}[ht]
\centering\small\begin{tabularx}{\linewidth}{l>{\raggedright\arraybackslash}Xrr}
\toprule
Context & Exact prompt & Choice & Included\\
\midrule
F1 & Please name one integer from the inclusive range of 1 to 30. & 15 & No\\
F2 & Choose a single whole number in the closed interval from 1 to 30. & 15 & No\\
F3 & I need one integer: choose any value from 1 through 30, endpoints included. & 23 & No\\
F4 & Which integer would you choose from 1 to 30, with both endpoints allowed? & Capped & No\\
F5 & Provide one whole number that is at least 1 and at most 30. & 25 & No\\
F6 & Select an integer from the set of whole numbers starting at 1 and ending at 30. & Capped & No\\
F7 & For this choice, give one integer in the range 1 through 30 inclusive. & 15 & No\\
F8 & You may choose any integer from 1 to 30 inclusive. Which one do you choose? & Capped & No\\
F9 & Name a single integer satisfying 1 \texttt{<=} n \texttt{<=} 30. & 5 & No\\
F10 & Make one numerical choice: an integer between 1 and 30, including the endpoints. & 17 & Yes\\
F11 & From the integers 1 through 30, choose one value for your answer. & 15 & No\\
F12 & What is your choice of one whole number in the inclusive interval 1 to 30? & 15 & No\\
\bottomrule
\end{tabularx}

\caption{All twelve fresh screening outcomes. ``Capped'' means no completed numerical choice within 80 tokens. All nine completed choices passed the format and alignment checks; eight were excluded because they were not 17. F10 passed the subsequent common-digit and write-cap checks. No reader/MLP intervention was collected for the other eleven contexts.}
\label{tab:fresh-screen}
\end{table}

For the eligible context, donor reader outputs were recaptured from the edited state with source-off values inserted at the pre-answer cache position in L32--33. Keys and other cache entries retained recipient values. The reader replacement acted at the query processing the common digit 1; the joint clamp held L34--35 MLP outputs at the matching no-reader recipient values. These hooks were removed before subsequent generation. Each reader effect was measured against the no-reader condition under the same clamp setting. The six core conditions crossed no reader, head 26 and complement with normal or joint-clamped processing. Four reference/identity conditions completed the ten-condition set.

\begin{table}[ht]
\centering\small\begin{tabular}{lrr}
\toprule
Incremental 7/5 effect & Normal & Joint L34--35 clamp\\
\midrule
Head 26 & $+0.168781$ & $+0.029083$\\
Complement & $+0.021667$ & $+0.018753$\\
Head minus complement & $+0.147114$ & $+0.010330$\\
\bottomrule
\end{tabular}

\caption{Fresh-context effects in F10, in nats. Its source-off and source-on 7/5 margins are $+5.852978$ and $+5.437187$. The frozen primary contrast subtracts the clamped head/complement gap from the normal gap, giving $+0.136784$ nat. The 93.0\% reduction describes this gap, not a fraction of natural preference mediated by the MLPs. All ten measured suffixes were \texttt{17**.}, completing the retained prefix \texttt{I choose **}.}
\label{tab:fresh-reader}
\end{table}

\FloatBarrier

Two historical context/dose combinations were replayed: the selected prompt, ``Pick one integer in the inclusive interval [1, 30].'', at dose 1.81884765625, and the reused holdout, ``Select a single integer from 1 through 30, inclusive.'', at dose 1.9097900390625 (1.05 times base). Each combination contributed ten conditions. These twenty anchor conditions are separate from the ten fresh conditions. All thirty measured conditions passed the numerical and identity checks; 110 potential fresh conditions were uncollected because of the eleven exclusions, not failed dispatches. The three capped baselines were reviewed as having no completed choice, including two in which the automatic parser extracted a numeral from explanatory text. The completion criterion had already excluded these responses. Automatic and reviewed labels remain separate in the \href{https://github.com/phunterlau/why17/blob/main/results/replication12_20261102/README.md}{results and audit}; the \href{https://github.com/phunterlau/why17/blob/main/experiments/replication12_20261102/protocol.json}{frozen protocol} retains all gates and condition definitions. The analysis reports this designed bank descriptively and makes no population recurrence-rate inference.
\FloatBarrier

\subsection{Composing earlier-state and terminal-output edits}\label{app:paired-example}
A separate Qwen3-4B study illustrates why two edits that select the same first digit need not select the same complete number. At the pre-answer position, an earlier L28:5371 edit changes later source-position caches, while a terminal L35:787 edit changes the current readout after those caches are computed. A suffix edit at L35:4145 acts when the first generated digit is processed. Table~\ref{tab:paired-example} uses reviewed responses from the two contexts, out of four requested, that met the format qualification; it combines historical anchor records with later measurements.

\begin{table}[ht]
\centering\small
\begin{tabular}{lrr}
\toprule
Program & Anchor & One new eligible wording\\
\midrule
No edit & 17 & 17\\
Terminal root edit, L35:787 & 24 & 23\\
Earlier state edit, L28:5371 & 27 & 27\\
Terminal suffix edit, L35:4145 & 14 & 14\\
Earlier state plus terminal suffix edits & 24 & 24\\
\bottomrule
\end{tabular}
\caption{A bounded illustration of composing interventions at different positions. Both root interventions select 2, but their continuations differ; the suffix intervention then changes the earlier edit's 27 to 24. These are completed, reviewed choices rather than predictions from a fixed-prefix diagnostic.}
\label{tab:paired-example}
\end{table}

The additive doses are $-94.687$, $+484.836$ and $+461.576$ for the earlier, root and suffix edits. The earlier write is 2.15--2.29 times the whole MLP-output norm, so sparse coordinate count does not imply a small perturbation. The two excluded contexts provide no intervention evidence. Labels come from an unblinded review of saved replies, and no independent adjudication was added. The example establishes composition under these conditions; it does not test natural recruitment, pair synergy or a link between these coordinates and the L29/head-26 pathway.

\FloatBarrier
\section{Distribution responses and contextual processing}\label{app:response-support}
\subsection{Context-dependent downstream responses}\label{app:context-response}
\subsubsection{Response and blocking measures}\label{app:response-methods}
For a common activation $z^\ast$, context $c$ and symmetric step $h$, write $\ell_c(z)$ for the nine centered log probabilities of leading digits 1 through 9. The recorded response is
\[
 K_c=\frac{\ell_c(z^\ast-h)-\ell_c(z^\ast+h)}{2h}.
\]
Its positive direction corresponds to decreasing the activation. The principal interval confirmation used $h=5$; the distinct post hoc 21--40 cancellation check used $h=2$ with a changed bank, prefix and operating point. This convention must be retained when comparing response signs.

The interval-blocking measurement compares directions after unit normalization, $\widetilde K=K/\lVert K\rVert$. For a blocked group $X$ and a context pair, it reports
\[
 R_X=1-\frac{\lVert\widetilde K_X(c_2)-\widetilde K_X(c_1)\rVert}
 {\lVert\widetilde K(c_2)-\widetilde K(c_1)\rVert}.
\]
The 0.69/0.60/0.85 values reported below average this measure across the registered contrasts. They do not denote fractions of answer probability or mediated behavior.

The instruction comparison retains response magnitude. For ordinary and instructed responses $K_O,K_s$, its normalized difference and blocking effect are
\[
 \eta_K=\left[\frac{\sum\lVert K_s-K_O\rVert^2}
 {\sum(\lVert K_s\rVert^2+\lVert K_O\rVert^2)/2}\right]^{1/2},
 \qquad
 R_G=1-\frac{\sum\lVert\Delta K_G\rVert^2}{\sum\lVert\Delta K\rVert^2},
\]
where $\Delta K=K_s-K_O$ and $G$ is the blocked group. Sums run over the matched instruction comparisons. These are the measures behind the 0.65/0.71/0.53 and 0.96/0.97/0.95 summaries. All response ratios require nonzero denominators. The documented definitions do not recover unavailable per-condition observations or the full nine-coefficient fit for the earlier predictor.

\subsubsection{The same write produces different responses}
A neuron's weight vector is fixed across prompts, but subsequent computation can transform its write differently, as self-repair studies show \cite{hydra,selfrepair}. Across seven intervals, mean pairwise cosines of direct write/readout responses were 0.996, 0.993 and 0.998, versus 0.66, 0.53 and 0.78 for full responses. The lower alignment indicates contextual changes in response direction. Responses use the symmetric perturbations and sign convention defined above.

At 21--40, leading 3 remains valid, yet the historical full response norm was about 0.02 versus 0.07 for the direct response. A separate post hoc FP32 check gave 0.019 versus 0.058 (Figure~\ref{fig:response-detail}A). The latter changed the bank, prefix, perturbation step and operating point; both configurations show cancellation without constituting a matched replication.

\subsubsection{Later MLP computation contributes to contextual variation}
We then held later MLP outputs at their recipient baseline to identify which computation contributes to the contextual difference. This removed, on average, 0.69, 0.60 and 0.85 of the response difference over three registered interval contrasts. These fractions describe response vectors; natural behavior mediation requires its own endpoint. The comparison lacked a matched source-neuron blocking panel, leaving source-specificity unresolved.

Instruction contrasts tested the same distinction at a common source activation. Instruction-dependent entropy steering provides a related distributional-control setting \cite{entropy}; here we measure how the recipient context changes the response to a fixed scalar edit. Ordinary and equal-probability requests yielded normalized response differences of 0.65, 0.71 and 0.53. Later-MLP blocking reduced their summed squared difference by 0.96, 0.97 and 0.95. The recipient context thus changed how the activation affected the output. The amount of natural distribution change carried by that activation is tested separately in Section~\ref{sec:natural-transfer}.

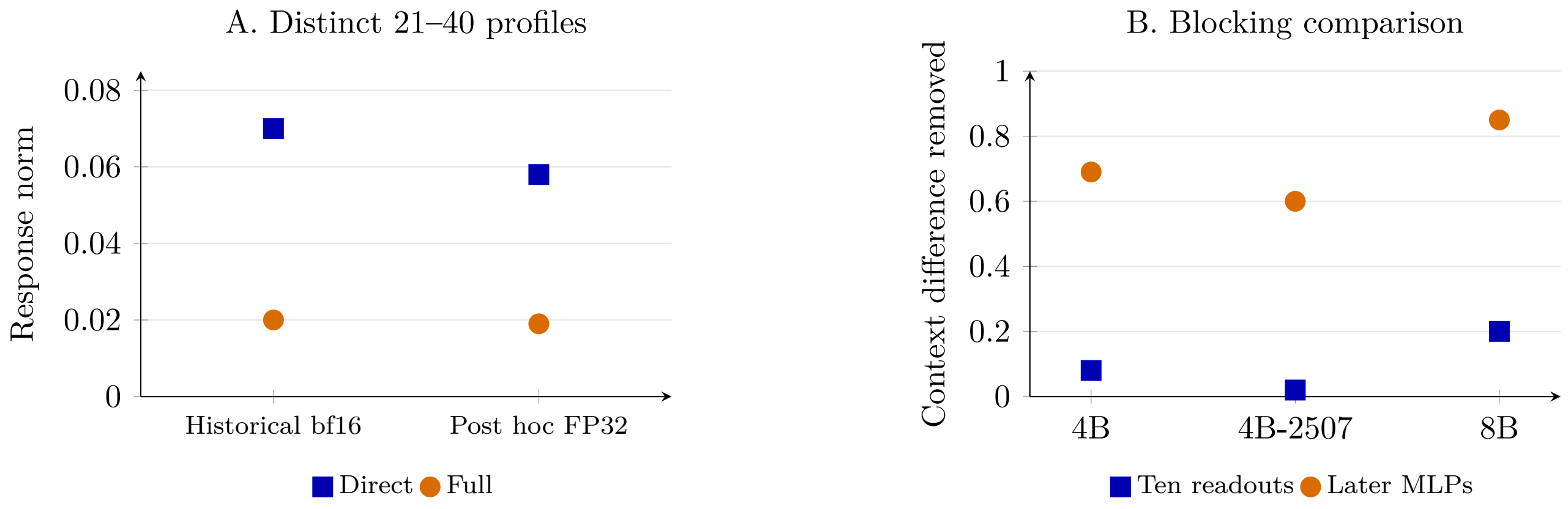
\begin{figure}[tb]
\centering\begin{tikzpicture}
\begin{axis}[width=.43\linewidth,height=5.1cm,axis lines=left,tick label style={font=\small},label style={font=\small},title style={font=\small},legend style={font=\scriptsize,draw=none},ymajorgrids,grid style={gray!20},xtick={1,2},xticklabels={Historical bf16,Post hoc FP32},x tick label style={font=\scriptsize},xmin=.5,xmax=2.5,ymin=0,ymax=.085,ylabel={Response norm},title={A. Distinct 21--40 profiles},scaled y ticks=false,yticklabel style={/pgf/number format/fixed},legend style={at={(.5,-.2)},anchor=north,legend columns=2}]
\addplot[blue!70!black,only marks,mark=square*,mark size=3pt] coordinates {(1,0.07) (2,0.058)};
\addlegendentry{Direct}
\addplot[orange!85!black,only marks,mark=*,mark size=3pt] coordinates {(1,0.02) (2,0.019)};
\addlegendentry{Full}

\end{axis}
\end{tikzpicture}\hfill
\begin{tikzpicture}
\begin{axis}[width=.43\linewidth,height=5.1cm,axis lines=left,tick label style={font=\small},label style={font=\small},title style={font=\small},legend style={font=\scriptsize,draw=none},ymajorgrids,grid style={gray!20},xtick={1,2,3},xticklabels={4B,4B-2507,8B},xmin=.7,xmax=3.3,ymin=0,ymax=1,ylabel={Context difference removed},title={B. Blocking comparison},legend style={at={(.5,-.2)},anchor=north,legend columns=2}]
\addplot[blue!70!black,only marks,mark=square*,mark size=3pt] coordinates {(1,0.08) (2,0.02) (3,0.2)};
\addlegendentry{Ten readouts}
\addplot[orange!85!black,only marks,mark=*,mark size=3pt] coordinates {(1,0.69) (2,0.6) (3,0.85)};
\addlegendentry{Later MLPs}

\end{axis}
\end{tikzpicture}
\caption{Downstream-response measurements. A: attenuation of a direct response at 21--40 in two distinct configurations; the FP32 check uses a different bank, prefix, dose and operating point. B: contextual response difference removed by blocking ten nominated readouts or the broader later MLPs, across the three checkpoints. The groups differ in size and coverage; this is not a matched-specificity test. No per-prompt observations or uncertainty are reconstructed from aggregate summaries.}
\label{fig:response-detail}
\end{figure}

Response size and contextual variation also depended differently on the tested components. Ten nominated late readout neurons supplied size fractions of 0.42, 0.31 and 0.18, while blocking them removed only 0.08, 0.02 and 0.20 of the contextual response difference. The broader later-MLP intervention removed substantially more (Figure~\ref{fig:response-detail}B). In the 4B instruction comparison, the best scalar gain was 0.74 and a reported 38\% of the response change concerned its shape. These response measurements show why a large immediate contribution alone gives an incomplete account of contextual processing.

These comparisons show contextual processing of the edit. Appendix~\ref{app:first-digit-accounting} resolves some later writes after source zeroing at 1--40 into unit contributions, without isolated mediation tests. Section~\ref{sec:path} follows a smaller 1--30 edit into a second-digit change. Sharing a source does not identify a common downstream circuit across these contexts, doses and endpoints.

\subsection{First-digit response accounting}\label{app:first-digit-accounting}
An earlier streamed-FP32 capture study used the pinned Qwen3-4B checkpoint on two historical punctuation forms of the 1--40 request. Zeroing L29:8172 only at the recorded first-numeral decision changed both complete replies from 23 to 17. The following saved accounting adds no validation cohort.

\begin{figure}[ht]
\centering\begin{tikzpicture}
\begin{axis}[width=.43\linewidth,height=5.1cm,axis lines=left,tick label style={font=\small},label style={font=\small},title style={font=\small},legend style={font=\scriptsize,draw=none},ymajorgrids,grid style={gray!20},xtick={1,2,3},xticklabels={Source,MLP32--35,Remaining},xmin=.7,xmax=3.3,ymin=0,ymax=100,ylabel={Share of total change (\%)},title={A. Numerator accounting},legend pos=north west]
\addplot[blue!70!black,mark=*,only marks] coordinates {(0.945,6.78667318) (1.945,87.3589341) (2.945,5.85436183)};
\addlegendentry{Wording A}
\addplot[orange!85!black,mark=square*,only marks] coordinates {(1.055,6.63545607) (2.055,86.5797412) (3.055,6.7847963)};
\addlegendentry{Wording B}

\end{axis}
\end{tikzpicture}\hfill
\begin{tikzpicture}
\begin{axis}[width=.43\linewidth,height=5.1cm,axis lines=left,tick label style={font=\small},label style={font=\small},title style={font=\small},legend style={font=\scriptsize,draw=none},ymajorgrids,grid style={gray!20},xtick={1,2},xticklabels={L34,L35},xmin=.7,xmax=2.3,ymin=0,ymax=100,ylabel={Share of layer write (\%)},title={B. Six reported units},legend pos=south west]
\addplot[blue!70!black,mark=*,only marks] coordinates {(0.945,78.4549177) (1.945,85.179873)};
\addlegendentry{Wording A}
\addplot[orange!85!black,mark=square*,only marks] coordinates {(1.055,78.5623494) (2.055,85.3284094)};
\addlegendentry{Wording B}

\end{axis}
\end{tikzpicture}
\caption{Two historical first-digit examples after zeroing the selected source at 1--40. A: signed numerator contributions divided by the total numerator change. ``Remaining'' includes the other recorded contributions. B: the six largest absolute unit contributions, summed with their signs and divided by the corresponding net layer contribution. The panels use different denominators; markers are offset horizontally for visibility. Both recorded replies change from 23 to 17; neither example is a new held-out replication. The accounting does not test serial causal links between units.}
\label{fig:first-digit-accounting}
\end{figure}
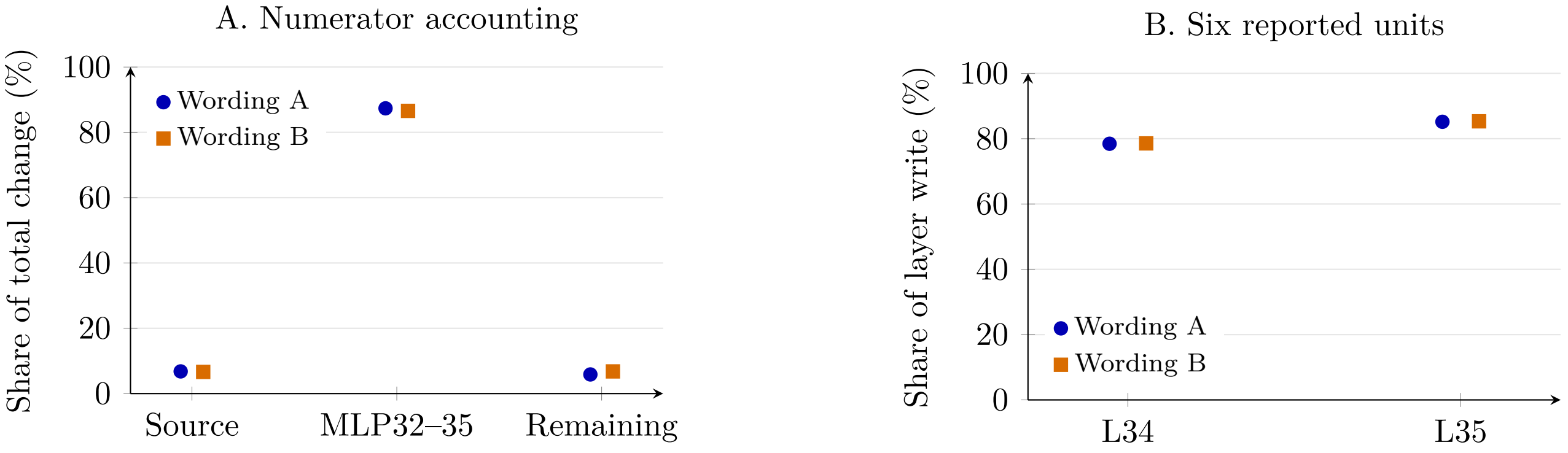

Projecting captured residual increments onto the gain-weighted 2-minus-1 readout before final normalization separates the source's removed write from later responses. In the first wording, the numerator changed by $-122.13$. The source write accounts for 6.79\% and the changed MLP32--35 writes for 87.36\% of that signed change. The second wording gives 6.64\% and 86.58\%, respectively (Figure~\ref{fig:first-digit-accounting}A). These are readout-accounting shares, with small closure residuals, rather than isolated effects of intervening on each later module.

Saved unit activations at L34 and L35 permit a finer decomposition. The six largest absolute contributions in each layer, selected retrospectively, sum to 78.45\% and 85.18\% of the corresponding net layer writes in the first wording. The second wording gives 78.56\% and 85.33\% (Figure~\ref{fig:first-digit-accounting}B). For example, L35:2165 changes from $-25.13$ to $-1.24$, while L35:672 changes from $-7.27$ to $-21.12$. Their contributions are the readout coupling multiplied by the activation change. This identifies responding units and their signed direct writes, without testing which unit recruits another or whether the selected set mediates the layer's effect. Signed contributions can cancel, so these fractions are not information shares or circuit-size estimates.

The intervention differs from the selected 17-to-15 dose study in Section~\ref{sec:path}, which changes continuation competition after the common first digit. The two studies share a source coordinate but do not establish two exclusive routes or a common downstream neuron circuit.

\subsection{Exact accounting of the saved candidate distributions}\label{app:branch-analysis}
Across eighteen positive source edits at 1--30, $C_a$ accounts for 59.1\% of the summed KL change, with per-context shares from 0.46\% to 99.9998\%. These shares depend jointly on the response and the probability distribution on which it acts. A substantial log-odds shift can contribute little KL when it affects an almost empty branch. The matched-write control has a higher within-branch share, 94.1\%, but much smaller summed movement (0.00269 versus 0.04918 nat). Shares alone therefore identify neither a different computational response nor source specificity.
We applied Equation~\ref{eq:branch-continuation} to all eighteen fixed-dose recipient baselines and their ninety self/source/control conditions, and to all eighteen dose conditions with arm-specific zero-dose baselines. Every vector contains the thirty candidates, log weights, normalized scores and candidate-event mass. The continuation component includes competition between stopping after one digit and emitting further digits; it is not exclusively a second-digit measure. The earlier twelve templates and six later extensions remain separate cohorts. No record was dropped or replaced, and the present reanalysis does not make these deterministic, dependent conditions independent replications.

\begin{table}[ht]
\centering\small
\begin{tabular}{lrrrr}
\toprule
Fixed-dose arm (18 contexts each) & $\sum B_a$ & $\sum C_a$ & $\sum C_a/\sum(B_a+C_a)$ & Eligible\\
\midrule
Positive source & 0.0200922 & 0.0290863 & 59.1\% & 18\\
Negative source & 0.0070568 & 0.0286929 & 80.3\% & 18\\
Positive control & 0.0001596 & 0.0025334 & 94.1\% & 15\\
Negative control & 0.0001890 & 0.0024630 & 92.9\% & 15\\
Self replacement & $1.68\times10^{-12}$ & $6.95\times10^{-11}$ & --- & 0\\
\bottomrule
\end{tabular}
\caption{Offline KL accounting in nats, with all eighteen contexts included in each sum. ``Eligible'' counts cells with total KL at least $10^{-8}$ nat, the prespecified numerical reporting threshold for per-cell fractions in this reanalysis. Absolute components are retained below it. The ratios describe these archived conditions rather than a population or source-specific property.}
\label{tab:branch-accounting}
\end{table}

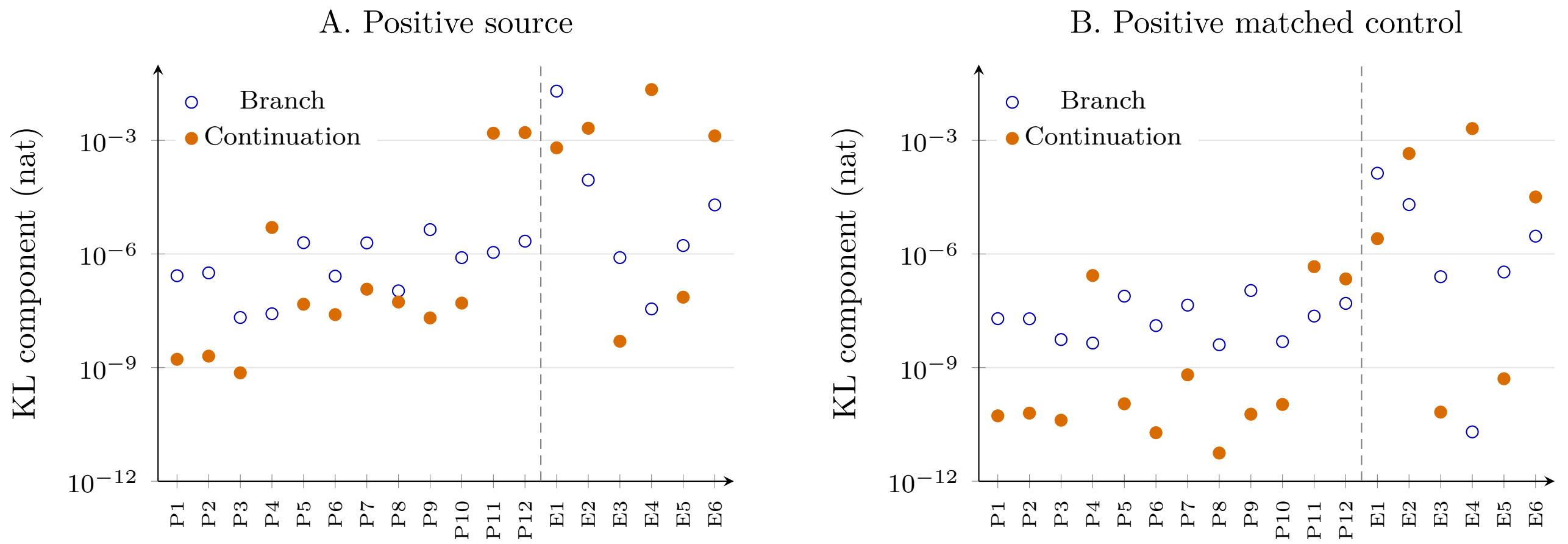
\begin{figure}[tb]
\centering\begin{tikzpicture}
\begin{axis}[width=.46\linewidth,height=6.1cm,axis lines=left,tick label style={font=\scriptsize},label style={font=\small},title style={font=\small},legend style={font=\scriptsize,draw=none},ymajorgrids,grid style={gray!20},xtick={1,2,3,4,5,6,7,8,9,10,11,12,13,14,15,16,17,18},xticklabels={P1,P2,P3,P4,P5,P6,P7,P8,P9,P10,P11,P12,E1,E2,E3,E4,E5,E6},x tick label style={font=\tiny,rotate=90,anchor=east},xmin=.4,xmax=18.6,ymode=log,ymin=1e-12,ymax=.1,ylabel={KL component (nat)},title={A. Positive source},legend pos=north west]
\addplot[blue!70!black,mark=o,mark size=1.8pt,only marks] coordinates {(1,2.66796674e-07) (2,3.16552899e-07) (3,2.1017076e-08) (4,2.63659093e-08) (5,1.99010314e-06) (6,2.59397491e-07) (7,1.96130339e-06) (8,1.06122789e-07) (9,4.3907473e-06) (10,8.00063045e-07) (11,1.10014291e-06) (12,2.17829073e-06) (13,0.019966932) (14,8.95697094e-05) (15,8.01441965e-07) (16,3.54117337e-08) (17,1.66975708e-06) (18,1.97447134e-05)};
\addlegendentry{Branch}
\addplot[orange!85!black,mark=*,mark size=1.8pt,only marks] coordinates {(1,1.65448284e-09) (2,2.00727321e-09) (3,7.27974117e-10) (4,5.0091576e-06) (5,4.70482512e-08) (6,2.50599965e-08) (7,1.17406928e-07) (8,5.38929542e-08) (9,2.04601413e-08) (10,5.04861304e-08) (11,0.00154469605) (12,0.00160327117) (13,0.000633934321) (14,0.00208773938) (15,4.95991431e-09) (16,0.0219007656) (17,7.22769261e-08) (18,0.00131052693)};
\addlegendentry{Continuation}
\addplot[gray,densely dashed] coordinates {(12.5,1e-12) (12.5,0.1)};

\end{axis}
\end{tikzpicture}\hfill
\begin{tikzpicture}
\begin{axis}[width=.46\linewidth,height=6.1cm,axis lines=left,tick label style={font=\scriptsize},label style={font=\small},title style={font=\small},legend style={font=\scriptsize,draw=none},ymajorgrids,grid style={gray!20},xtick={1,2,3,4,5,6,7,8,9,10,11,12,13,14,15,16,17,18},xticklabels={P1,P2,P3,P4,P5,P6,P7,P8,P9,P10,P11,P12,E1,E2,E3,E4,E5,E6},x tick label style={font=\tiny,rotate=90,anchor=east},xmin=.4,xmax=18.6,ymode=log,ymin=1e-12,ymax=.1,ylabel={KL component (nat)},title={B. Positive matched control},legend pos=north west]
\addplot[blue!70!black,mark=o,mark size=1.8pt,only marks] coordinates {(1,1.9596529e-08) (2,1.94504823e-08) (3,5.50478702e-09) (4,4.44262384e-09) (5,7.68926282e-08) (6,1.28470753e-08) (7,4.45993443e-08) (8,4.01203113e-09) (9,1.08254374e-07) (10,4.84358396e-09) (11,2.29183534e-08) (12,4.96180844e-08) (13,0.000135616045) (14,2.01097561e-05) (15,2.50768576e-07) (16,2.00619577e-11) (17,3.34289292e-07) (18,2.95037335e-06)};
\addlegendentry{Branch}
\addplot[orange!85!black,mark=*,mark size=1.8pt,only marks] coordinates {(1,5.32193917e-11) (2,6.27273661e-11) (3,4.07905843e-11) (4,2.69996745e-07) (5,1.10864072e-10) (6,1.90610719e-11) (7,6.44777517e-10) (8,5.53960451e-12) (9,5.88653685e-11) (10,1.06358297e-10) (11,4.62758848e-07) (12,2.19497871e-07) (13,2.53294093e-06) (14,0.000449101111) (15,6.67160978e-11) (16,0.00204886491) (17,5.07760333e-10) (18,3.19478832e-05)};
\addlegendentry{Continuation}
\addplot[gray,densely dashed] coordinates {(12.5,1e-12) (12.5,0.1)};

\end{axis}
\end{tikzpicture}
\caption{Absolute branch and continuation KL components for the eighteen positive-dose L29:8172 interventions and their L29:8369 matched controls, on a common logarithmic scale. P1--P12 are the earlier templates; E1--E6 are later extensions. E1 and E2 initially favor 23 and 15; the other candidate modes are 17. E4 is the subsequently selected switching prompt. Source and matched-write control use separate panels. Small components remain visible instead of being converted into large-looking percentages. No prompt-level uncertainty is estimated.}
\label{fig:branch-accounting}
\end{figure}

For the positive source, the twelve earlier templates contribute 0.003167 nat in total, 99.6\% within branches; the six extensions contribute 0.046012 nat, 56.4\% within branches. Figure~\ref{fig:branch-accounting} shows the large differences in absolute movement that these aggregate shares conceal. The two largest source movements contribute 86.4\% of the total. In the first extension, branch movement contributes 0.019967 of 0.020601 nat while mode 23 is preserved. In the selected fourth extension, continuation movement contributes 0.02190077 of 0.02190080 nat while the candidate mode changes from 17 to 15. The dose study separately verifies the generated answer change.

All eight nonzero target doses in the selected prompt have continuation shares above 99.9997\%; all eight control doses exceed 99.99999\% with smaller movements. At unit dose their total KL values are 0.021901 and 0.002049 nat. The control opposes the source's 17/15 movement and never changes the generated answer. A high continuation share locates output redistribution without identifying a carrier or source specificity.

We reconstructed normalized scores from saved log weights and independently checked the KL expansion and the branch-oracle residual. Maximum absolute closure errors were below $8\times10^{-17}$ nat; direct KL calculations agreed within $2.3\times10^{-16}$ nat. Self replacements had maximum KL $4.38\times10^{-11}$ nat and TV $4.59\times10^{-6}$; the ratio threshold avoids interpreting fractions at that scale. Candidate-event mass ranged from 0.754205 to 1.000000160 across the 126 input vectors. The small numerical excesses above one are retained, not clipped. Paired mass comparisons identify one material source change: E1 falls from 0.835538 to 0.754205, a decrease of 8.13 percentage points, while its matched control increases mass by 0.005775. The other seventeen source changes are below $10^{-5}$ in absolute value; the selected E4 changes by about $-8.8\times10^{-8}$. Excluded event mass is not necessarily invalid-answer mass. These masses refer to canonical digits followed by any nondigit; saved scores cannot be converted to probabilities of an exact numeral followed by an end-of-sequence (EOS) token without additional suffix probabilities.

\paragraph{The selected continuation crossing.}
At the base dose, exact candidate accounting gives $B_a=3.54\times10^{-8}$ and $C_a=0.0219008$ nat. Almost all measured KL is within branches, concentrated on 17 and 15. Yet leading-2 versus leading-1 log-odds change by $-0.609$ nat, larger in magnitude than the $-0.420$ nat change between 17 and 15. Branch 2 initially holds only $2.84\times10^{-7}$ candidate mass, while 17 and 15 both have substantial probability. The tiny branch KL therefore does not mean a negligible relative branch response. The unchanged first digit and measured second-digit change, rather than the KL share alone, motivate the cache study.

\subsection{Response profiles, prediction losses and controls}\label{app:profile-prediction}

For prompt $j$, define $t_j(n)=\log q_{a,j}(n)-\log q_{0,j}(n)$. Then $q_{a,j}(n)\propto q_{0,j}(n)\exp t_j(n)$, with both baseline and response allowed to vary. Figure~\ref{fig:response-geometry}A centers each profile over thirty candidates and displays the all-eighteen median. The retrospective prediction instead adds the coordinate-wise median from the other seventeen profiles to the recipient's baseline log probabilities and renormalizes. It is not a frozen predictor test.

Sixteen profiles have Pearson correlation at least 0.96 with their leave-one-out median, including the selected E4 (0.979 over thirty numbers; 0.956 within 10--19). E1 and E2 give 0.895 and 0.819 over all candidates. Centering each profile before the nonlinear median retains the same sixteen above 0.96. These checks do not identify a rank-one mechanism or a common internal route; unlikely candidates receive the same weight as likely ones. The leave-one-out continuation-share rank correlation is 0.988, with all eighteen comparisons on the observed side of 50\%. For E4, however, the estimate retains 17, leaves a $+0.109570$ margin, and yields only 44.3\% of observed total KL. The observed 17/15 change is $-0.419716$ nat versus the panel median $-0.281745$.

For a minor branch of baseline mass $p$ and finite log-odds change $D$, the leading small-$p$ KL term is $p[D e^D-e^D+1]$. This approximation explains how a substantial relative shift can contribute little KL: the selected branch-2 shift acts on mass $2.84\times10^{-7}$, whereas 17 and 15 both carry substantial mass. Branch log-odds responses can themselves depend on within-branch mixtures; they are not baseline-independent coefficients.

\paragraph{Oracle ceiling on branch-only error reduction.}
If $\widehat q_a(n)=\widehat p_a(d(n))r_0(n\mid d(n))$ preserves the baseline conditionals, then
\[
 \KL(q_a\Vert\widehat q_a)=C_a+\KL(p_a\Vert\widehat p_a),\qquad
 R_{\mathrm{pred}}=\frac{\sum_a B_a-\sum_a\KL(p_a\Vert\widehat p_a)}{\sum_a(B_a+C_a)}.
\]
The branch share of summed KL is thus an oracle ceiling for this score on the same evaluation set. The eighteen-prompt archive cannot supply that ceiling for the missing 21-interval evaluation. For the historical predictor in Appendix~\ref{app:historical-prediction}, Equation~\ref{eq:branch-predictor}, adding one constant to every $\beta_d$ leaves the normalized predictor unchanged; a recovered fit should specify its additive-gauge convention. Neither the complete fit nor that convention is reconstructed here.

\FloatBarrier
\paragraph{Prediction losses and matched controls.}
In the selected prompt, the raw-median estimate moves 44.3\% as far from baseline in KL; that movement ratio is not prediction-error reduction.

Within 10--19, source/control median leave-one-out correlations are 0.970/0.959, while their within-prompt alignment is opposing (median $-0.874$). The source's median centered amplitude is 7.36 times larger. Across all thirty candidates, correlations are 0.983/0.779, and the source has a larger leading-2/leading-1 shift. Table~\ref{tab:profile-controls} gives all four sites. Shared output shape alone establishes neither source specificity nor a common internal route.

Appendix~\ref{app:zeroing-comparison} separately reports the cross-interval sign reversal and concentration of pooled KL, with dose and prefix changes retained. The shared shapes above concern wordings at one interval and dose; Appendix~\ref{app:context-response} examines downstream response across different interval and instruction settings.
For each leave-one-out estimate, let $D_j=\KL(q_{a,j}\Vert q_{0,j})$ and $L_j=\KL(q_{a,j}\Vert\widehat q_{a,j})$. The pooled error reduction is $1-\sum_jL_j/\sum_jD_j$; prompt-balanced summaries use the individual reductions $1-L_j/D_j$. Table~\ref{tab:profile-errors} retains all eighteen losses and total variation errors. The raw-profile median gives pooled reduction 0.89032, mean per-prompt reduction 0.92223 and median 0.96254. Every $D_j$ exceeds $2.17\times10^{-8}$ nat, although tiny absolute movements limit the interpretation of some relative improvements. Sixty-digit arithmetic agreed with the computed KL values within $4.17\times10^{-17}$ nat.

\begin{table}[ht]
\centering\small\begin{tabular}{lrrrr}
\toprule
Context & Baseline loss $D_j$ & Prediction loss $L_j$ & Reduction & TV error\\
\midrule
P1 & $2.68\times10^{-7}$ & $4.43\times10^{-9}$ & 98.35\% & $6.69\times10^{-8}$\\
P2 & $3.19\times10^{-7}$ & $1.71\times10^{-9}$ & 99.46\% & $4.01\times10^{-8}$\\
P3 & $2.17\times10^{-8}$ & $2.19\times10^{-9}$ & 89.93\% & $5.07\times10^{-8}$\\
P4 & $5.04\times10^{-6}$ & $1.35\times10^{-6}$ & 73.20\% & $6.37\times10^{-6}$\\
P5 & $2.04\times10^{-6}$ & $2\times10^{-8}$ & 99.02\% & $6.79\times10^{-7}$\\
P6 & $2.84\times10^{-7}$ & $5.43\times10^{-9}$ & 98.09\% & $1.39\times10^{-7}$\\
P7 & $2.08\times10^{-6}$ & $1.75\times10^{-9}$ & 99.92\% & $1.85\times10^{-7}$\\
P8 & $1.6\times10^{-7}$ & $5.89\times10^{-9}$ & 96.32\% & $8.56\times10^{-8}$\\
P9 & $4.41\times10^{-6}$ & $9.26\times10^{-8}$ & 97.90\% & $2.66\times10^{-6}$\\
P10 & $8.51\times10^{-7}$ & $1.67\times10^{-8}$ & 98.04\% & $3.04\times10^{-7}$\\
P11 & $0.00155$ & $0.000268$ & 82.66\% & $0.00218$\\
P12 & $0.00161$ & $0.000187$ & 88.33\% & $0.0021$\\
\midrule
E1 & $0.0206$ & $0.0019$ & 90.78\% & $0.015$\\
E2 & $0.00218$ & $0.000354$ & 83.75\% & $0.0021$\\
E3 & $8.06\times10^{-7}$ & $1.69\times10^{-8}$ & 97.90\% & $1.08\times10^{-6}$\\
E4 & $0.0219$ & $0.00243$ & 88.89\% & $0.0349$\\
E5 & $1.74\times10^{-6}$ & $6.64\times10^{-8}$ & 96.19\% & $3.42\times10^{-6}$\\
E6 & $0.00133$ & $0.000249$ & 81.29\% & $0.00244$\\
\bottomrule
\end{tabular}

\caption{Retrospective leave-one-out prediction for each saved positive source edit. Losses are in nats; TV compares predicted with observed candidate-event distributions. P1--P12 are historical templates and E1--E6 extensions. These related contexts are not independent replications. The selected E4 is the only mode mismatch; an unchanged-baseline predictor also matches the other seventeen modes.}
\label{tab:profile-errors}
\end{table}

Pooled reductions are 0.85570 on the twelve historical prompts and 0.89270 on the six extensions; mean TV errors are 0.000357 and 0.009064. The all-context mean TV is 0.003260, whereas its median is $1.87\times10^{-6}$. Thus a small typical absolute error does not characterize the selected prompt, whose TV error is 0.034864. Its prediction loss is 0.002434055 nat versus baseline loss 0.021900801 nat, giving 88.886\% reduction despite retaining the wrong winner.

Centering each thirty-number response before the coordinate-wise median changes the pooled reduction to 0.87744. The selected margin becomes $+0.122669$ nat and its reduction 86.703\%; it still retains 17. The nonlinear median therefore need not preserve the prediction under different centering conventions. A separate descriptive split estimates the profile from only the twelve historical templates and evaluates the six extensions: pooled reductions are 0.89272 without centering and 0.88103 with it. These analyses were chosen after the outputs were available and are not prospective validation.

The branch-only comparator applies median donor changes in branch log mass to each recipient, preserving its baseline conditionals and using the same leave-one-out donors. Pooled reduction is 0.38541, median per-prompt reduction 0.90736 and mean TV 0.008437. This descriptive comparator is neither an optimized fit nor the historical nine-coefficient model. The pooled/median difference reflects concentrated movement.

\begin{table}[ht]
\centering\small\begin{tabular}{lrrr}
\toprule
Site & All-30 correlation & 10--19 correlation & 10--19 amplitude\\
\midrule
L29:8172 & 0.9826 & 0.9705 & 0.20478\\
L29:8369 (control) & 0.7794 & 0.9592 & 0.02781\\
L34:2836 & 0.9994 & 0.9943 & 0.01483\\
L34:1261 (control) & 0.8695 & 0.9652 & 0.00352\\
\bottomrule
\end{tabular}

\caption{Positive-dose output responses at four separately edited sites. Correlation entries are medians over eighteen per-context Pearson correlations with the leave-one-out coordinate-wise median response. The 10--19 columns restrict each response to that integer range. Amplitude is the median root mean square of the response centered within 10--19, in nats. This amplitude statistic is distinct from PP ranking aggregation. Equal write sizes apply within each source/control pair, not across layers.}
\label{tab:profile-controls}
\end{table}

Within-prompt teen-profile correlations with L29:8172 have medians $-0.874$ for its control, $-0.944$ for L34:2836 and $-0.833$ for the L34 control. Median relative writes are 0.04793 for both L29 sites and 0.01294 for both L34 sites. These selected and matched sites supply comparators for output shape; they do not constitute a population sample or identify a common internal path.

\subsection{Zeroing across two interval cohorts}\label{app:zeroing-comparison}

A separate FP32 replay retained twelve discovery-template distributions per interval before and after zeroing the same source. Table~\ref{tab:zeroing-regimes} applies Equation~\ref{eq:branch-continuation}, summing twelve edit-versus-unedited comparisons per row.

The pooled 1--30 percentage is dominated by two templates. The median per-template branch share is 87.7\%, with branch KL larger in nine of twelve comparisons; the largest two cells hold 98.2\% of total KL. At 1--40 the corresponding values are 99.991\%, twelve of twelve, and 30.3\%. Table~\ref{tab:zeroing-template} reports each comparison. Thus summed and typical-template shares describe different aspects of the same records.

\begin{table}[ht]
\centering
\small
\begin{tabular}{lrrrr}
\toprule
Interval & Total KL & Branch KL & Continuation KL & Branch share\\
\midrule
1--30 & 0.025055 & 0.0000767 & 0.024979 & 0.306\%\\
1--40 & 40.395118 & 40.236408 & 0.158710 & 99.607\%\\
\bottomrule
\end{tabular}
\caption{Archived zeroing comparisons, with twelve discovery templates per interval. Values are summed KL components in nats and their branch share; these are previously used contexts with context-dependent zeroing doses.}
\label{tab:zeroing-regimes}
\end{table}

Zeroing changes the source by its context-dependent baseline value. At 1--30 the unedited log-probability vectors match the fixed-positive-dose archive exactly across the twelve common templates. Their response shapes correlate 0.9935--0.9983 over thirty numbers and above 0.997 within the teens. Fitted response scales track dose ratios (correlation 0.957; median scale/dose ratio 0.936), consistent with approximate scaling at these two doses. Two dose points do not establish general linearity. The 17/15 contrast decreases in every 1--30 template (median $-1.470$ nat) but increases in every 1--40 template (median $+1.462$ nat). Cross-interval response correlations within the teens have median $-0.101$. This is a change in relative response, beyond a difference in KL weighting, but dose and prefix also change with interval.

Scores reconstructed from log weights match saved normalized values within $10^{-12}$, and chain-rule residuals are below $10^{-12}$ nat. These reused cohorts do not recover the older 21-interval predictor. The positive-dose comparisons use unedited recipients as baselines; self-replacement effects remain separately reported.

\begin{table}[ht]
\centering\small
\begin{tabular}{lrrrrrr}
\toprule
 & \multicolumn{3}{c}{1--30} & \multicolumn{3}{c}{1--40}\\
\cmidrule(lr){2-4}\cmidrule(lr){5-7}
Template & Total KL & Branch share & $\Delta_{17/15}$ & Total KL & Branch share & $\Delta_{17/15}$\\
\midrule
P1 & $1.48\times10^{-6}$ & 99.333\% & $-0.879$ & $6.22$ & 100.000\% & $+1.693$\\
P2 & $1.72\times10^{-6}$ & 99.190\% & $-0.692$ & $6$ & 100.000\% & $+1.479$\\
P3 & $6.02\times10^{-7}$ & 92.147\% & $-1.786$ & $4.12$ & 99.999\% & $+0.977$\\
P4 & $0.000396$ & 0.066\% & $-1.556$ & $0.639$ & 81.475\% & $+2.033$\\
P5 & $1.17\times10^{-5}$ & 95.969\% & $-1.170$ & $1.67$ & 99.992\% & $+1.048$\\
P6 & $1.43\times10^{-6}$ & 84.983\% & $-1.335$ & $3.56$ & 99.998\% & $+0.860$\\
P7 & $1.33\times10^{-5}$ & 90.159\% & $-1.533$ & $2.4$ & 99.977\% & $+1.607$\\
P8 & $9.24\times10^{-7}$ & 60.332\% & $-1.507$ & $2.96$ & 99.983\% & $+1.581$\\
P9 & $2.36\times10^{-5}$ & 98.867\% & $-1.434$ & $2.87$ & 99.989\% & $+1.544$\\
P10 & $4.23\times10^{-6}$ & 85.219\% & $-1.148$ & $4.8$ & 99.995\% & $+1.110$\\
P11 & $0.0111$ & 0.060\% & $-2.270$ & $2.71$ & 99.306\% & $+1.445$\\
P12 & $0.0135$ & 0.105\% & $-2.261$ & $2.44$ & 99.198\% & $+1.196$\\
\bottomrule
\end{tabular}

\caption{Per-template zeroing results, rounded for display. Total KL is in nats; branch share is the branch component divided by that template's total. $\Delta_{17/15}$ is the change in candidate log-odds of 17 against 15. P1--P12 match the historical template labels in Figure~\ref{fig:branch-accounting}. These are discovery contexts with context-dependent doses.}
\label{tab:zeroing-template}
\end{table}

\subsection{Historical wording and interval evidence}\label{app:historical-prediction}
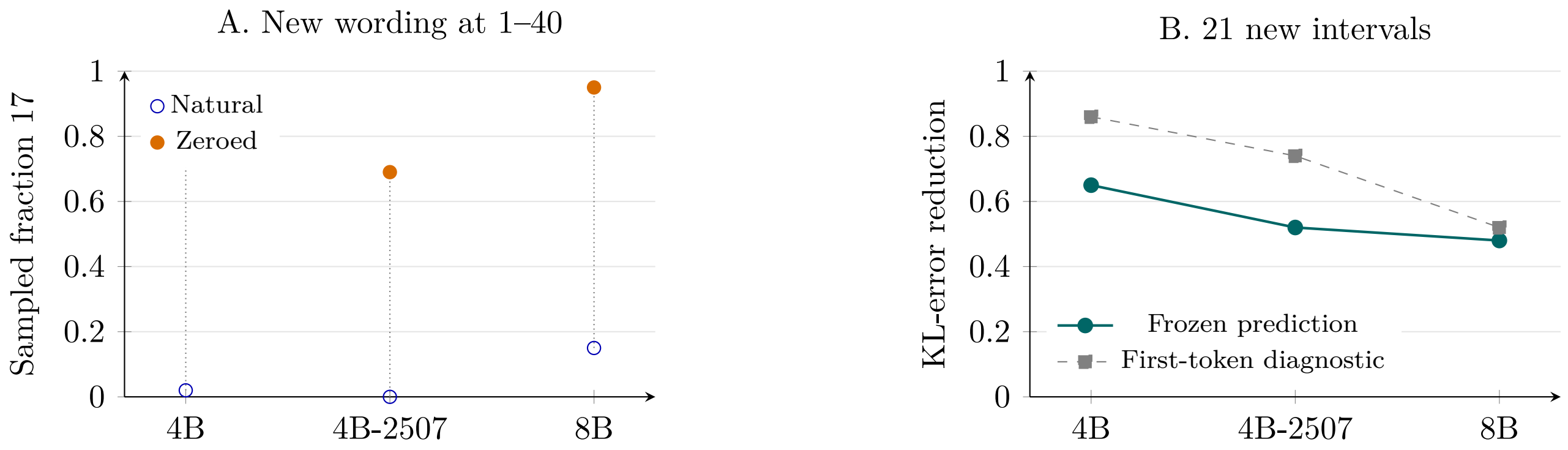
\begin{figure}[tb]
\centering\begin{tikzpicture}
\begin{axis}[width=.43\linewidth,height=5.1cm,axis lines=left,tick label style={font=\small},label style={font=\small},title style={font=\small},legend style={font=\scriptsize,draw=none},ymajorgrids,grid style={gray!20},xtick={1,2,3},xticklabels={4B,4B-2507,8B},xmin=.7,xmax=3.3,ymin=0,ymax=1,ylabel={Sampled fraction 17},title={A. New wording at 1--40},legend pos=north west]
\addplot[blue!70!black,mark=o,only marks] coordinates {(1,0.02) (2,0) (3,0.15)};
\addlegendentry{Natural}
\addplot[orange!85!black,mark=*,only marks] coordinates {(1,0.77) (2,0.69) (3,0.95)};
\addlegendentry{Zeroed}
\addplot[gray,densely dotted] coordinates {(1,0.02) (1,0.77)};
\addplot[gray,densely dotted] coordinates {(2,0) (2,0.69)};
\addplot[gray,densely dotted] coordinates {(3,0.15) (3,0.95)};

\end{axis}
\end{tikzpicture}\hfill
\begin{tikzpicture}
\begin{axis}[width=.43\linewidth,height=5.1cm,axis lines=left,tick label style={font=\small},label style={font=\small},title style={font=\small},legend style={font=\scriptsize,draw=none},ymajorgrids,grid style={gray!20},xtick={1,2,3},xticklabels={4B,4B-2507,8B},xmin=.7,xmax=3.3,ymin=0,ymax=1,ylabel={KL-error reduction},title={B. 21 new intervals},legend pos=south west]
\addplot[teal!80!black,mark=*,thick] coordinates {(1,0.65) (2,0.52) (3,0.48)};
\addlegendentry{Frozen prediction}
\addplot[gray,mark=square*,dashed] coordinates {(1,0.86) (2,0.74) (3,0.52)};
\addlegendentry{First-token diagnostic}

\end{axis}
\end{tikzpicture}
\caption{Validation and prediction for the selected coordinates. A: sampled fractions on eligible new wordings (16, 17 and 20 prompts); connecting segments pair the two conditions within each checkpoint. Bootstrap uncertainty on the increases is given in the text. B: the frozen predictor and a separate first-token diagnostic on 21 new intervals. Model order is fixed throughout. No point-level data or uncertainty absent from the retained historical summaries is reconstructed.}
\label{fig:foundation}
\end{figure}

\paragraph{Historical interval prediction.}
To test whether the intervention has structure beyond its effect on 17, we fitted a predictor that assigns an activation-dependent shift to each leading-digit branch:
\begin{equation}
 \widehat q_a(n)\ \propto\ q_0(n)\exp\{\Delta u\,\beta_{d(n)}\},
\label{eq:branch-predictor}
\end{equation}
where $\Delta u$ measures movement toward a more negative source activation. Nine branch coefficients were fitted before evaluation. Exponential reweighting of baseline probabilities also appears in analyses of steering strength \cite{strength}. The empirical question here is whether a frozen leading-digit profile predicts effects of an internal neuron edit on new intervals. Prior probing and intervention work motivates examining digit structure \cite{base10}; these coefficients describe output response and do not by themselves identify a semantic representation of digits.

For a frozen prediction $\widehat q_a$ of a clamped distribution $q_a$, the historical evaluation used
\begin{equation}
 R_{\mathrm{pred}}=1-\frac{\sum_a\KL(q_a\Vert\widehat q_a)}{\sum_a\KL(q_a\Vert q_0)}.
 \label{eq:pred}
\end{equation}
The resulting score measures relative Kullback--Leibler (KL) prediction-error reduction.

The earlier FP32 summaries report $R_{\mathrm{pred}}=0.65,0.52,0.48$ on 21 new intervals. Cross-entropy improvements relative to no change are 0.17, 0.17 and 0.15 nat; alternative numerical-value, relative-position and favored-number-rank accounts improve by at most 0.04 nat. The full coefficient fit, exact evaluation grid and per-condition observations are unavailable in the current evidence packet. These values therefore remain documented aggregate results rather than a newly audited predictor evaluation.

The historical fits favor leading 3 and 2 over 1 as activation becomes more negative. Zeroing at 1--40 removes that competing contribution, allowing leading 1 to gain while the remaining computation favors 17. This interpretation is specific to the interval fits. It assigns no fixed number to the coordinate and does not exclude contextual or within-branch changes.

A separate retrospective calculation replaced only the first-token probabilities using the observed intervention. It accounted for 0.86, 0.74 and 0.52 of the KL movement (Figure~\ref{fig:foundation}B). Its access to the observed intervention distinguishes this decomposition from the prospective predictor. Its residual differs from the frozen predictor's error, which can also reflect imperfect branch coefficients. Raw first-token probabilities need not equal the branch marginals of the candidate-conditioned distribution, because the scoring event also includes the continuation and termination. Its pooled FP32 share was 0.698, below the internally recorded 0.70 criterion; this failed threshold is separate from the fitted predictor's error reduction. We keep the diagnostic separate from the exact accounting in Appendix~\ref{app:branch-analysis}.

\FloatBarrier
\section{Natural-context transfer diagnostics}\label{app:transfer-diagnostics}
\begin{figure}[tb]
\centering\begin{tikzpicture}
\begin{axis}[width=.43\linewidth,height=5.1cm,axis lines=left,tick label style={font=\small},label style={font=\small},title style={font=\small},legend style={font=\scriptsize,draw=none},ymajorgrids,grid style={gray!20},xtick={1,2,3},xticklabels={4B,4B-2507,8B},xmin=.7,xmax=3.3,ymin=0,ymax=1.1,ylabel={Projection fit summary},title={A. Scaled-zeroing diagnostic},legend pos=south west]
\addplot[teal!80!black,only marks,mark=*,mark size=3pt] coordinates {(1,0.91) (2,0.89) (3,0.98)};
\addlegendentry{Correlation}
\addplot[orange!85!black,only marks,mark=square*,mark size=3pt] coordinates {(1,0.76) (2,0.82) (3,0.82)};
\addlegendentry{Slope}

\end{axis}
\end{tikzpicture}\hfill
\begin{tikzpicture}
\begin{axis}[width=.43\linewidth,height=5.1cm,axis lines=left,tick label style={font=\small},label style={font=\small},title style={font=\small},legend style={font=\scriptsize,draw=none},ymajorgrids,grid style={gray!20},xtick={1,2,3},xticklabels={4B,4B-2507,8B},xmin=.7,xmax=3.3,ymin=-.025,ymax=.075,ylabel={Projected transfer},title={B. Cancellation in the average},scaled y ticks=false,yticklabel style={/pgf/number format/fixed},legend pos=north east]
\addplot[blue!70!black,only marks,mark=square*,mark size=3pt] coordinates {(1,0.056) (2,0.05) (3,0.023)};
\addlegendentry{Mean absolute}
\addplot[orange!85!black,only marks,mark=*,mark size=3pt] coordinates {(1,-0.012) (2,0.018) (3,0.004)};
\addlegendentry{Signed mean}
\addplot[gray,dashed] coordinates {(0.7,0) (3.3,0)};

\end{axis}
\end{tikzpicture}
\caption{Post hoc diagnostics of instruction swaps. A: correlation and calibration slope of the scaled-zeroing projection approximation. B: mean absolute and signed projected transfer. Opposing effects reduce the signed average. The panels show aggregates for the specified projection readout. JS-based donorward progress and independent validation require different measurements; per-pair observations are not reconstructed here.}
\label{fig:weak-transfer}
\end{figure}
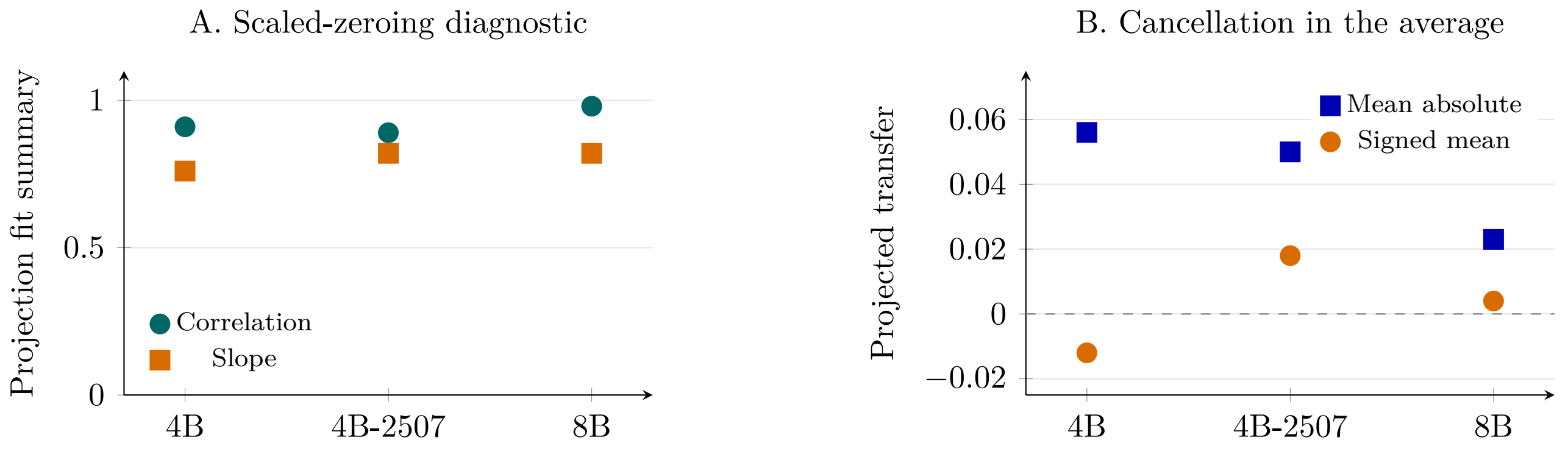

\paragraph{Interpreting donorward progress.}
Negative progress indicates movement away from the donor. Because JS divergence is bounded, this score is not a linear fraction of log-odds movement. For example, in a two-answer distribution, moving log-odds from $-16$ to $-4$ covers half the distance to a donor at $+8$, but gives only $F=0.0654$. This mathematical example motivates reporting continuous displacement and TV alongside the normalized score. The pooled measure weights contrasts through their divergences, and its interpretation depends on the scored distribution. First-digit, candidate-answer and fixed-leading-digit distributions answer different questions about generation. We use matched comparisons where available and retain each readout and cohort when reporting transfer, following methodological work on activation patching \cite{patching}.
\paragraph{Activation changes and local sensitivity.}
Post hoc analyses examined why the activation swaps produced small average transfers. For instruction pairs, an approximation scaled the zeroing effect by the natural activation displacement relative to the zeroing distance. On a signed projection readout, correlations with measured transfers were 0.91, 0.89 and 0.98, with calibration slopes of 0.76, 0.82 and 0.82 (Figure~\ref{fig:weak-transfer}). The approximation retained the signs among the largest five percent of transfers. Small displacements and limited effects restricted individual changes; opposing signs further reduced the average. Mean absolute projected transfers were 0.056, 0.050 and 0.023, compared with signed means of $-0.012$, $+0.018$ and $+0.004$.

History produced substantial activation changes but a weak response at the second answer. The earlier value changed the natural activation by 4.9 and 7.2 units in 4B and 8B, while the neuron's local effect was only about one percent of the history contrast. A local-derivative approximation tracked projected transfers with reported correlations and slopes near 1.00. This combination explains how a large change of activation can still transfer little of the measured output contrast.

Both diagnostics use local intervention information retrospectively; whole-distribution errors were larger. They differ from the historical interval predictor and JS progress in Equation~\ref{eq:transfer}. Sixteen matched control neurons also had small instruction-transfer effects. These readout-specific checks relate weak average transfer to displacement, sensitivity and cancellation.

\FloatBarrier
\section{Requested sampling laws and mathematical limits}\label{app:distribution}
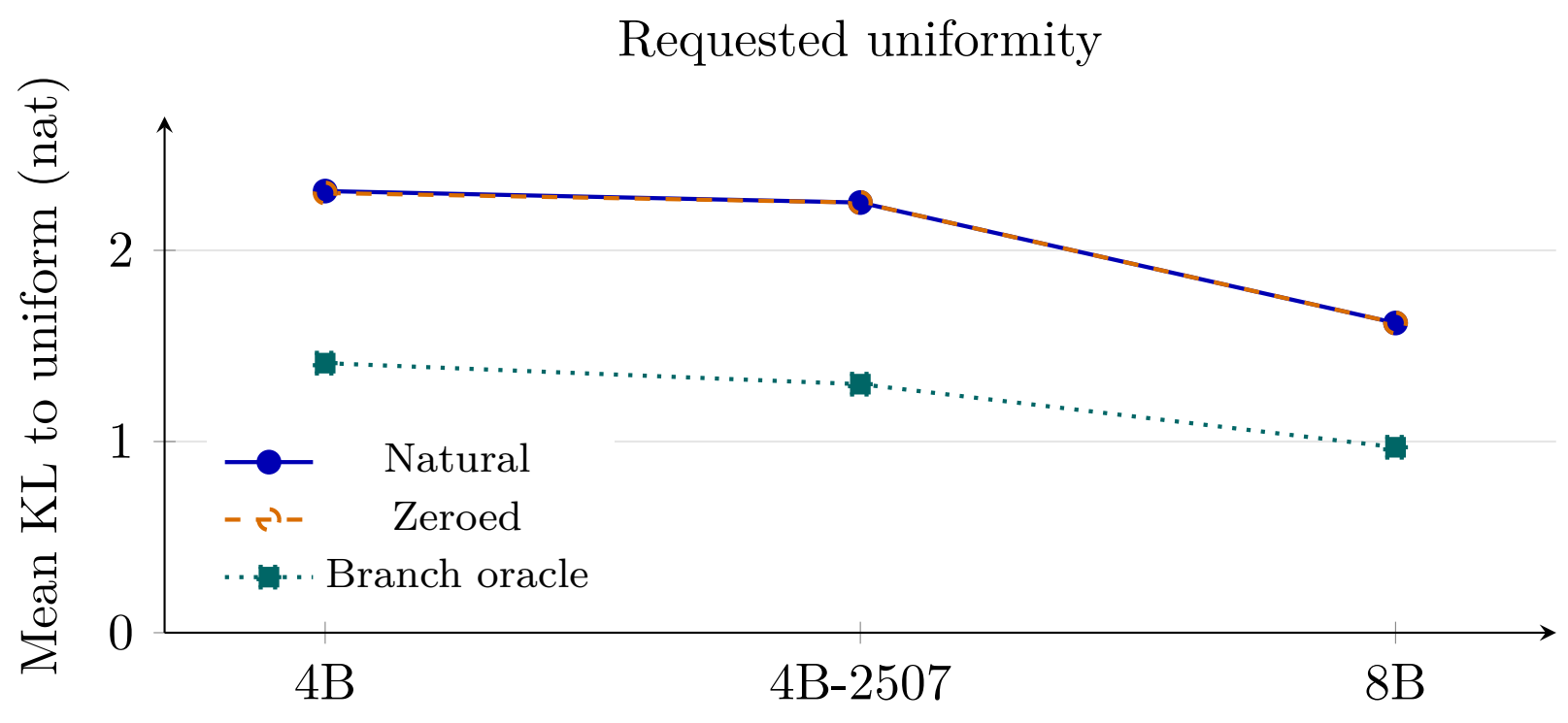
\begin{figure}[htbp]
\centering\begin{tikzpicture}
\begin{axis}[width=.65\linewidth,height=5.1cm,axis lines=left,tick label style={font=\small},label style={font=\small},title style={font=\small},legend style={font=\scriptsize,draw=none},ymajorgrids,grid style={gray!20},xtick={1,2,3},xticklabels={4B,4B-2507,8B},xmin=.7,xmax=3.3,ymin=0,ymax=2.7,ylabel={Mean KL to uniform (nat)},title={Requested uniformity},legend pos=south west]
\addplot[blue!70!black,mark=*,thick] coordinates {(1,2.31) (2,2.25) (3,1.62)};
\addlegendentry{Natural}
\addplot[orange!85!black,mark=o,dashed,thick] coordinates {(1,2.3) (2,2.25) (3,1.62)};
\addlegendentry{Zeroed}
\addplot[teal!80!black,mark=square*,dotted,thick] coordinates {(1,1.41) (2,1.3) (3,0.97)};
\addlegendentry{Branch oracle}

\end{axis}
\end{tikzpicture}
\caption{Mean KL to uniform in the separate instruction-bank analysis. The oracle changes branch weights while holding conditional distributions fixed. Lines connect model summaries for readability; they do not fit a scaling law. This sampling-law comparison is separate from the matched natural-transfer assay in Figure~\ref{fig:limits}.}
\label{fig:sampling-limits}
\end{figure}
A valid chosen number and a requested sampling distribution impose different requirements. Earlier studies distinguish independent requests from autoregressive lists and show that the protocol affects the measured distribution \cite{renda,dice}. An unconstrained request permits any in-range choice, while explicit uniform or independent sampling instructions concern the distribution over replies. In the instruction-bank analysis, zeroing changed numerical preference but barely changed mean KL to uniform: 2.31 to 2.30, 2.25 to 2.25, and 1.62 to 1.62 nat. Mean absolute changes of 0.18, 0.21 and 0.10 nat show cancellation in these small net effects. This retrospective analysis used a reference prefix, six intervals, twelve confirmation families and three request forms.

\subsection{Uniformity under fixed within-branch laws}
A branch-only controller also has a calculable limitation. If $q(d,r)=a_d r_d(r)$ and only $a$ is adjustable, entropy maximization gives
\begin{equation}
 \min_a\KL(q\Vert U)=\log N-\log\sum_d\exp H(r_d).
 \label{eq:floor}
\end{equation}
The mean oracle floors were 1.41, 1.30 and 0.97 nat in this bank (Figure~\ref{fig:sampling-limits}). The bound describes control restricted to branch weights with fixed conditional continuations. The actual neuron also affected continuation competition, as the selected second-digit experiment shows.

For $q(d,r)=a_d r_d(r)$, expanding $-\sum q\log q$ gives
$H(q)=H(a)+\sum_d a_dH(r_d)$. Write $Z=\sum_d e^{H(r_d)}$ and $a_d^*=e^{H(r_d)}/Z$. Then
\[
H(q)=\log Z-\KL(a\Vert a^*).
\]
Maximizing over branch weights yields Equation~\ref{eq:floor}. This derivation assumes fixed within-branch conditional laws and a uniform target over $N$ candidates.

\subsection{History dependence and branch reweighting}\label{app:math}
Response-history studies report changes in marginal bias \cite{bscore}, and West and Potts \cite{west} show why improved marginal balance can coexist with non-independent repetition patterns. We therefore separate marginal imbalance from history dependence. For second draws with balanced forced earlier values $f$, let $\bar q=\E_f q_f$. Then
\begin{equation}
 \E_f\KL(q_f\Vert U)=\underbrace{\KL(\bar q\Vert U)}_{B}
 +\underbrace{\E_f\KL(q_f\Vert\bar q)}_{D}.
\end{equation}
In the source/framing experiments, explicit independent/uniform wording lowered $B+D$ in all 52 comparisons, with scored mass 0.98--1.00. The comparisons shared stems and comprised forty source cells and twelve framing cells. These dependencies limit their interpretation as repeated evidence. The measured reduction establishes an output-level effect whose internal mechanism remains unresolved.

The chain rule gives $D=I(F;D_1)+I(F;R\mid D_1)$, with branch and continuation contributions under the observed distribution. Changing branch probabilities can also change the continuation term by reweighting branches and conditional mixtures, even if each law given $(F,D_1)$ stays fixed. The derivation below makes this dependence explicit. The observed continuation share therefore describes the measured distribution; it supplies no invariant lower bound on history dependence under arbitrary branch control.

\Needspace{8\baselineskip}

For history dependence, the joint distribution also includes the earlier value $F$. Expanding the log ratio through $\bar q$ gives
\[
\E_f\KL(q_f\Vert U)=\E_f\KL(q_f\Vert\bar q)+\KL(\bar q\Vert U).
\]
The dependence term is mutual information under the balanced forced-value design. Its continuation component is
\[
 I(F;R\mid D_1)=\sum_d p(d)\sum_f p(f\mid d)
 \KL\bigl(q(r\mid f,d)\Vert q(r\mid d)\bigr).
\]
Changing only $p(d\mid f)$ can alter the outside weights, the posterior $p(f\mid d)$ and the mixture $q(r\mid d)$. Fixed $q(r\mid f,d)$ therefore does not imply a fixed conditional-information term.

For an explicit counterexample, let $F$ be a fair bit and choose branch A with probability $t$ independently of $F$. In A, set $R=F$; in branch B, draw $R$ as an independent fair bit. Every within-branch law is fixed, but $I(F;D_1)=0$ and $I(F;R\mid D_1)=t\log2$. Reweighting $t$ from 0.8 to 0.2 changes the latter from 0.554518 to 0.138629 nat. This constructed mathematical example illustrates the identity independently of language-model observations. It rules out treating a measured continuation-information share as invariant under arbitrary branch control.


\begin{thebibliography}{99}
\small
\setlength{\itemsep}{1pt}
\bibitem{tools} T. Blankenstein, J. Yu, Z. Li, V. Plachouras, S. Sengupta, P. Torr, Y. Gal, A. Paren and A. Bibi. BiasBusters: Uncovering and Mitigating Tool Selection Bias in Large Language Models. ICLR, 2026. \url{https://arxiv.org/abs/2510.00307v2}.
\bibitem{languages} L. Twist, M. Harman, D. Syme, J. Noppen, H. Yannakoudakis, D. Nauck and J. M. Zhang. A Study of LLMs' Preferences for Libraries and Programming Languages. Findings of ACL, 2026. \url{https://arxiv.org/abs/2503.17181v4}.
\bibitem{numbers} J. Coronado-Bl\'azquez. Deterministic or probabilistic? The psychology of LLMs as random number generators. 2025. \url{https://arxiv.org/abs/2502.19965}.
\bibitem{stochastic} X. Gu, S. De, M. Titsias, L. Markeeva, P. Veli\v{c}kovi\'c and R. Pascanu. The Illusion of Stochasticity in LLMs. 2026. \url{https://arxiv.org/abs/2604.06543}.
\bibitem{qwen} Qwen Team. Qwen3 Technical Report. 2025. \url{https://arxiv.org/abs/2505.09388}.
\bibitem{vig} J. Vig et al. Causal Mediation Analysis for Interpreting Neural NLP: The Case of Gender Bias. 2020. \url{https://arxiv.org/abs/2004.12265}.
\bibitem{patching} F. Zhang and N. Nanda. Towards Best Practices of Activation Patching in Language Models: Metrics and Methods. 2023. \url{https://arxiv.org/abs/2309.16042}.
\bibitem{illusion} A. Makelov, G. Lange and N. Nanda. Is This the Subspace You Are Looking for? An Interpretability Illusion for Subspace Activation Patching. 2023 arXiv version. \url{https://arxiv.org/abs/2311.17030}.
\bibitem{hydra} T. McGrath, M. Rahtz, J. Kramar, V. Mikulik and S. Legg. The Hydra Effect: Emergent Self-repair in Language Model Computations. 2023. \url{https://arxiv.org/abs/2307.15771}.
\bibitem{pp} H. Liu, T.-L. Li and Y. Wu. Perturbation Probing: A Two-Pass-per-Prompt Diagnostic for FFN Behavioral Circuits in Aligned LLMs. 2026. \url{https://arxiv.org/abs/2604.27401}.
\bibitem{control} H. Liu. Leverage Is Not Reach: A Control-Window Law for Single-Neuron Steering in Language Models. 2026. \url{https://arxiv.org/abs/2606.19831}.
\bibitem{sparse} A. Arora, Z. Wu, J. Steinhardt and S. Schwettmann. Language Model Circuits Are Sparse in the Neuron Basis. ICML, PMLR 306:3712--3753, 2026. \url{https://proceedings.mlr.press/v306/arora26c.html}.
\bibitem{benford} J. Shao, Y. Lu and J. Yang. Benford's Curse: Tracing Digit Bias to Numerical Hallucination in LLMs. NeurIPS, 2025. \url{https://arxiv.org/abs/2506.01734v2}.
\bibitem{modular} T. Baeumel, D. Gurgurov, Y. Al Ghussin, J. van Genabith and S. Ostermann. Modular Arithmetic: Language Models Solve Math Digit by Digit. Findings of IJCNLP-AACL, pp. 1380--1409, 2025. \url{https://aclanthology.org/2025.findings-ijcnlp.86/}.
\bibitem{base10} A. A. Levy and M. Geva. Language Models Encode Numbers Using Digit Representations in Base 10. NAACL, pp. 385--395, 2025. \url{https://aclanthology.org/2025.naacl-short.33/}.
\bibitem{ansari} S. Ansari. 3, 7, 14, 17, 42: Instruction Tuning Installs and Sharpens Number-Choice Heuristics in Language Models. arXiv preprint, 2026. \url{https://arxiv.org/abs/2610.36178}.
\bibitem{featureuse} T. Che and Y. Li. Does the Model Use the Feature? Separating Steering from Mechanism in LLMs. arXiv preprint, 2026. \url{https://arxiv.org/abs/2610.07270}.
\bibitem{strength} M. Taimeskhanov, S. Vaiter and D. Garreau. Towards Understanding Steering Strength. ICML, PMLR 306:118084--118133, 2026. \url{https://proceedings.mlr.press/v306/taimeskhanov26a.html}.
\bibitem{selfrepair} C. Rushing and N. Nanda. Explorations of Self-Repair in Language Models. ICML, PMLR 235:42836--42855, 2024. \url{https://proceedings.mlr.press/v235/rushing24a.html}.
\bibitem{cachesteer} M. Belitsky, D. J. Kopiczko, M. Dorkenwald, M. J. Mirza, J. R. Glass, C. G. M. Snoek and Y. M. Asano. KV Cache Steering for Controlling Frozen LLMs. arXiv preprint, 2025. \url{https://arxiv.org/abs/2507.08799v2}.
\bibitem{entropy} L. Baroni, K. Ayonrinde, S. Feng and P. Radmard. Language Models Can Coarsely Modulate Output Entropy Under Instruction. AI4GOOD Workshop at ICML, 2026. \url{https://openreview.net/forum?id=SWBs3iyjs2}.
\bibitem{renda} A. Renda, A. Hopkins and M. Carbin. Can LLMs Generate Random Numbers? Evaluating LLM Sampling in Controlled Domains. ICML Workshop on Sampling and Optimization in Discrete Space (SODS), 2023. \url{https://people.csail.mit.edu/renda/llm-sampling-paper}.
\bibitem{west} P. West and C. Potts. Base Models Beat Aligned Models at Randomness and Creativity. COLM, 2025. \url{https://arxiv.org/abs/2505.00047}.
\bibitem{atp} J. Kram\'ar, T. Lieberum, R. Shah and N. Nanda. AtP*: An Efficient and Scalable Method for Localizing LLM Behaviour to Components. arXiv preprint, 2024. \url{https://arxiv.org/abs/2403.00745}.
\bibitem{ioi} K. Wang, A. Variengien, A. Conmy, B. Shlegeris and J. Steinhardt. Interpretability in the Wild: A Circuit for Indirect Object Identification in GPT-2 Small. ICLR, 2023. \url{https://arxiv.org/abs/2211.00593}.
\bibitem{heuristics} Y. Nikankin, A. Reusch, A. Mueller and Y. Belinkov. Arithmetic Without Algorithms: Language Models Solve Math with a Bag of Heuristics. ICLR, 2025. \url{https://proceedings.iclr.cc/paper_files/paper/2025/hash/8c5f30296296d2ae402ebbd09aaa9c12-Abstract-Conference.html}.
\bibitem{bscore} A. Vo, M. R. Taesiri, D. Kim and A. T. Nguyen. B-score: Detecting Biases in Large Language Models Using Response History. ICML, PMLR 267:61652--61688, 2025. \url{https://proceedings.mlr.press/v267/vo25a.html}.
\bibitem{dice} M. Zhao, Y. Du and M. Wang. Large Language Models Are Bad Dice Players: LLMs Struggle to Generate Random Numbers from Statistical Distributions. ACL, pp. 22942--22959, 2026. \url{https://aclanthology.org/2026.acl-long.1051/}.
\bibitem{ffn} M. Geva, A. Caciularu, K. Wang and Y. Goldberg. Transformer Feed-Forward Layers Build Predictions by Promoting Concepts in the Vocabulary Space. EMNLP, pp. 30--45, 2022. \url{https://aclanthology.org/2022.emnlp-main.3/}.
\end{thebibliography}
\end{document}